\documentclass[10pt,journal,compsoc]{IEEEtran}

\usepackage{enumitem}
\usepackage{comment}
\usepackage{color}

\ifCLASSOPTIONcompsoc
  \usepackage[nocompress]{cite}
\else
  \usepackage{cite}
\fi
\ifCLASSINFOpdf
  \usepackage[pdftex]{graphicx}
\else
\fi
\usepackage{amssymb}
\usepackage[cmex10]{amsmath}
\ifCLASSOPTIONcompsoc
  \usepackage[caption=false,font=footnotesize,labelfont=sf,textfont=sf]{subfig}
\else
  \usepackage[caption=false,font=footnotesize]{subfig}
\fi
\usepackage[export]{adjustbox}
\usepackage{booktabs}
\usepackage{multirow}
\usepackage{color}
\usepackage{hueifang}
\usepackage[mediumspace,mediumqspace,squaren]{SIunits}

\DeclareMathOperator*{\argmin}{arg\,min}

\begin{document}
%
\title{Supervised Learning of Semantics-Preserving Hash via Deep Convolutional Neural Networks}
%
%
%
%

\author{Huei-Fang~Yang,
        Kevin~Lin,
        and~Chu-Song~Chen
\IEEEcompsocitemizethanks{\IEEEcompsocthanksitem H.-F.~Yang is with Research Center for Information Technology Innovation, Academia Sinica, Taipei, Taiwan.  
E-mail: hfyang@iis.sinica.edu.tw 
\IEEEcompsocthanksitem K.~Lin is with the Dept.~Electrical Engineering, University of Washington, Seattle, WA, USA.  
E-mail: kvlin@uw.edu
\IEEEcompsocthanksitem C.-S.~Chen is with Institute of Information Science, Academia Sinica, Taipei, Taiwan.  
E-mail: song@iis.sinica.edu.tw
}
}

%
%

\markboth{IEEE Transactions on Pattern Analysis and Machine Intelligence}%
{Yang \MakeLowercase{\textit{et al.}}: Supervised Learning of Semantics-Preserving Hash via Deep Convolutional Neural Networks}
%



\IEEEtitleabstractindextext{%
\begin{abstract}
This paper presents a simple yet effective supervised deep hash approach that constructs binary hash codes from labeled data for large-scale image search. We assume that the semantic labels are governed by several latent attributes with each attribute \textit{on} or \textit{off}, and classification relies on these attributes. Based on this assumption, our approach, dubbed supervised semantics-preserving deep hashing (SSDH), constructs hash functions as a latent layer in a deep network and the binary codes are learned by minimizing an objective function defined over classification error and other desirable hash codes properties. With this design, SSDH has a nice characteristic that classification and retrieval are unified in a single learning model. Moreover, SSDH performs joint learning of image representations, hash codes, and classification in a point-wised manner, and thus is scalable to large-scale datasets. SSDH is simple and can be realized by a slight enhancement of an existing deep architecture for classification; yet it is effective and outperforms other hashing approaches on several benchmarks and large datasets. Compared with state-of-the-art approaches, SSDH achieves higher retrieval accuracy, while the classification performance is not sacrificed.
\end{abstract}

\begin{IEEEkeywords}
Image retrieval, supervised hashing, binary codes, deep learning, convolutional neural networks.
\end{IEEEkeywords}}

\maketitle

\IEEEdisplaynontitleabstractindextext

%
\IEEEpeerreviewmaketitle

\ifCLASSOPTIONcompsoc
\IEEEraisesectionheading{\section{Introduction}\label{sec:introduction}}
\else
\section{Introduction}
\label{sec:introduction}
\fi

%
%
%
%
\IEEEPARstart{S}{emantic} search is important in content-based image retrieval (CBIR).
Hashing methods that construct similarity-preserving binary codes for efficient image search have received great attention in CBIR~\cite{gong:pami13,he:kdd10,kulis:pami12}. 
The key principle in devising the hash functions is to map images of similar content to similar binary codes, which amounts to mapping the high-dimensional visual data into a low-dimensional Hamming (binary) space.
Having done so, one can perform an approximate nearest-neighbor (ANN) search by simply calculating the Hamming distance between binary vectors, an operation that can be done extremely fast.

Recently, learning-based hash approaches have become popular as they leverage training samples in code construction.
The learned binary codes are more efficient than the ones by locality sensitive hashing (LSH)~\cite{andoni:focs06} 
that maps similar images to the same bucket with high probability through random projections, makes no use of training data, and thus requires longer codes to attain high search accuracy.
Among various learning-based approaches, supervised hashing that exploits the supervised information (e.g., pairwised similarities or triple-wised rankings devised by data labels) during the hash
function construction can learn binary codes better capturing the semantic structure of data.
Though supervised hashing approaches yield promising performance, many of the recent techniques employ pairs or triplets of the training samples in the learning phase and thus require a long computation time and a high storage cost for training.
They are suitable for small-scale datasets but would be impractical when the data size becomes large.

Recent advances reveal that deep convolutional neural networks (CNNs) are capable of learning rich mid-level representations effective for image classification, object detection, and semantic segmentation~\cite{krizhevsky:nips12,
simonyan:iclr15,szegedy:cvpr15,girshick:cvpr14,sermanet:iclr14,long:cvpr15}.
The deep CNN architectures trained on a huge dataset of numerous categories (e.g., ImageNet~\cite{russakovsky:ijcv15}) can be transferred to new domains by employing them as feature extractors on other tasks including \textcolor{black}{recognition~\cite{donahue:icml14,oquab:cvpr14} and retrieval~\cite{razavian:cvprws14,babenko:eccv14}}, which provide better performance than handcrafted features such as GIST~\cite{oliva:ijcv01} and HOG~\cite{dalal:cvpr05}.
Moreover, the CNN parameters pre-trained on a large-scale dataset can be transferred and further \emph{fine-tuned} to perform a new task in another domain (such as PASCAL VOC~\cite{everingham:ijcv10}, Caltech-101~\cite{fei-fei:cvprw04}, \textcolor{black}{Oxford buildings~\cite{philbin:cvpr07}}) and capture more favorable semantic information of images~\cite{girshick:pami16,chatfield:bmvc14}.

The success of deep CNN on classification and detection tasks is encouraging.
It reveals that fine-tuning a CNN pre-trained on a large-scale and diverse-category dataset provides a fairly promising way for domain adaptation and transfer learning.
For image retrieval, a question worthy of study thus arises:
Beyond classification, is the ``pre-train + fine-tune" scheme also capable of learning binary hash codes for efficient retrieval?
Besides, if it is, how to modify the architecture of a pre-trained CNN to this end?

In this paper, to answer the question and 
enable efficient training with large-scale data, we take advantage of deep learning and propose the supervised semantics-preserving deep hashing (SSDH) for learning binary codes from labeled 
images.
The idea 
of SSDH is unsophisticated and innovated, where we assume that image labels can be implicitly represented by a set of latent attributes (\ie, binary codes) and the classification is dependent on these attributes.
Based on this idea, we construct the hash functions as a hidden layer between image representations and classification outputs in a CNN, and the binary codes 
are learned by minimizing an objective function defined over classification error and other desired properties on the binary codes.
This design yields a simple and effective 
network that unifies classification and retrieval in a single learning process and 
enforces semantically similar images to have similar binary codes.

Moreover, to make the outputs of each hidden node close to $0$ or $1$ and the resulting hash codes more separated, we impose additional constraints on the learning objective to make each hash bit carry as much information as possible and more discriminative.
During network learning, we transfer the parameters of the pre-trained network to SSDH and fine-tune SSDH on the target domains for efficient retrieval.
An overview of our approach is given 
in Figure~\ref{fig:deep-hashing}.

Our method can exploit existing well-performed deep convolution networks and provide an easy way to enhance them. 
Only a lightweight modification has been made on the architecture to achieve simultaneous classification and retrieval, and we show that the classification performance will not be sacrificed when our modification is applied.
Main contributions of this paper include:

\noindent \textbf{Unifying retrieval and classification}: SSDH is a supervised hash approach that takes advantage of deep learning, unifies classification and retrieval in a single learning model, and jointly learns representations, hash functions, and classification from image data.

\noindent \textbf{Scalable deep hash}: SSDH performs learning in a point-wised manner, and thereby requires neither pairs nor triplets of training inputs.
This characteristic makes it more scalable to large-scale data learning and retrieval.

\noindent \textbf{Lightweight deep hash}: SSDH is established upon the effective deep architecture and parameters pre-trained for classification; it can benefit from supervised deep transfer learning and is easily realizable by a slight enhancement of an existing deep classification network.

We conduct extensive experiments on several benchmarks and also some large collections of more than $1$ million images.
Experimental results show that our method is simple but powerful, and can easily generate more favorable results than existing state-of-the-art hash function learning methods. 
This paper is an extended version of~\cite{lin:cvprw15,lin:icmr15}.


\vspace*{-7pt}

\section{Background}\label{sec:related-work}

\subsection{Learning-based Hash}
Learning-based hash algorithms construct hash codes by leveraging the training data and are expected to overcome the limitations of data-independent methods in the LSH family~\cite{andoni:focs06, 
raginsky:nips09}.
The learning-based approaches can be grouped into three categories according to the degree of supervised information of labeled data used: unsupervised, semi-supervised, and supervised methods.

Unsupervised algorithms~\cite{gong:pami13,kulis:pami12,liu:icml11,weiss:nips08} use unlabeled data for code construction and try to preserve the similarity between data examples in the original space (\eg, the Euclidean space).
Representative methods 
include spectral hashing (SH)~\cite{weiss:nips08}, kernelized locality-sensitive hashing (KLSH)~\cite{kulis:pami12}, and iterative quantization (ITQ)~\cite{gong:pami13}.

Semi-supervised algorithms~\cite{mu:cvpr10,wang:pami12,wang:eccv14} use information from both labeled and unlabeled samples for learning hash functions.
For example, the 
SSH~\cite{wang:pami12} minimizes the empirical error on the pairwise labeled data (\eg, similar and dissimilar data pairs) and maximizes the variance of hash codes. 
The semi-supervised tag hashing (SSTH)~\cite{wang:eccv14} models the correlation between the hash codes and the class labels in a supervised manner and preserves the similarity between image examples in an unsupervised manner.

Supervised hashing approaches~\cite{kulis:nips09,lin:iccv13,lin:cvpr14,liu:cvpr12,norouzi:icml11,norouzi:nips12,
shen:cvpr15,wang:iccv13} aim to fully take advantage of the supervised information of labeled data for learning more efficient binary representations, therefore attaining higher search accuracy than the unsupervised and the semi-supervised approaches.
Utilizing pairwise relations between data samples, binary reconstructive embedding (BRE)~\cite{kulis:nips09} minimizes the squared error between the original Euclidean distances and the Hamming distances of binary codes, and the same/different labels information can be integrated in the training scheme for supervision.
Minimal loss hashing (MLH)~\cite{norouzi:icml11} minimizes the empirical loss for code construction.
Ranking-based methods~\cite{norouzi:nips12,wang:iccv13} that leverage the ranking information from a set of triplets have also been proposed.
Methods that rely on pairs or triplets of image samples for training generally need a high storage cost and are infeasible for large datasets.
Learning binary codes in a point-wised manner would be a better alternative for the scalability of hash.
Point-wise methods use the provided label information to guide the learning of hash functions.
Iterative quantization with canonical correlation analysis (CCA-ITQ)~\cite{gong:pami13} applies CCA with label information for dimensionality reduction and then performs binarization through minimizing the quantization error.
The supervised discrete hashing (SDH)~\cite{shen:cvpr15} formulates the learning of hash codes in terms of classification in order to learn binary codes optimal for classification.
While SDH and ours share similar spirits on coupling hash code learning and classification, SDH decomposes the hashing learning into sub-problems and needs a careful choice of loss function for classification to make the entire optimization efficient and scalable.
Our formulation on the deep networks simplifies the optimization process and is naturally scalable to large-scale datasets.


\begin{figure}
	\centering
	\includegraphics[width=1.1\columnwidth]{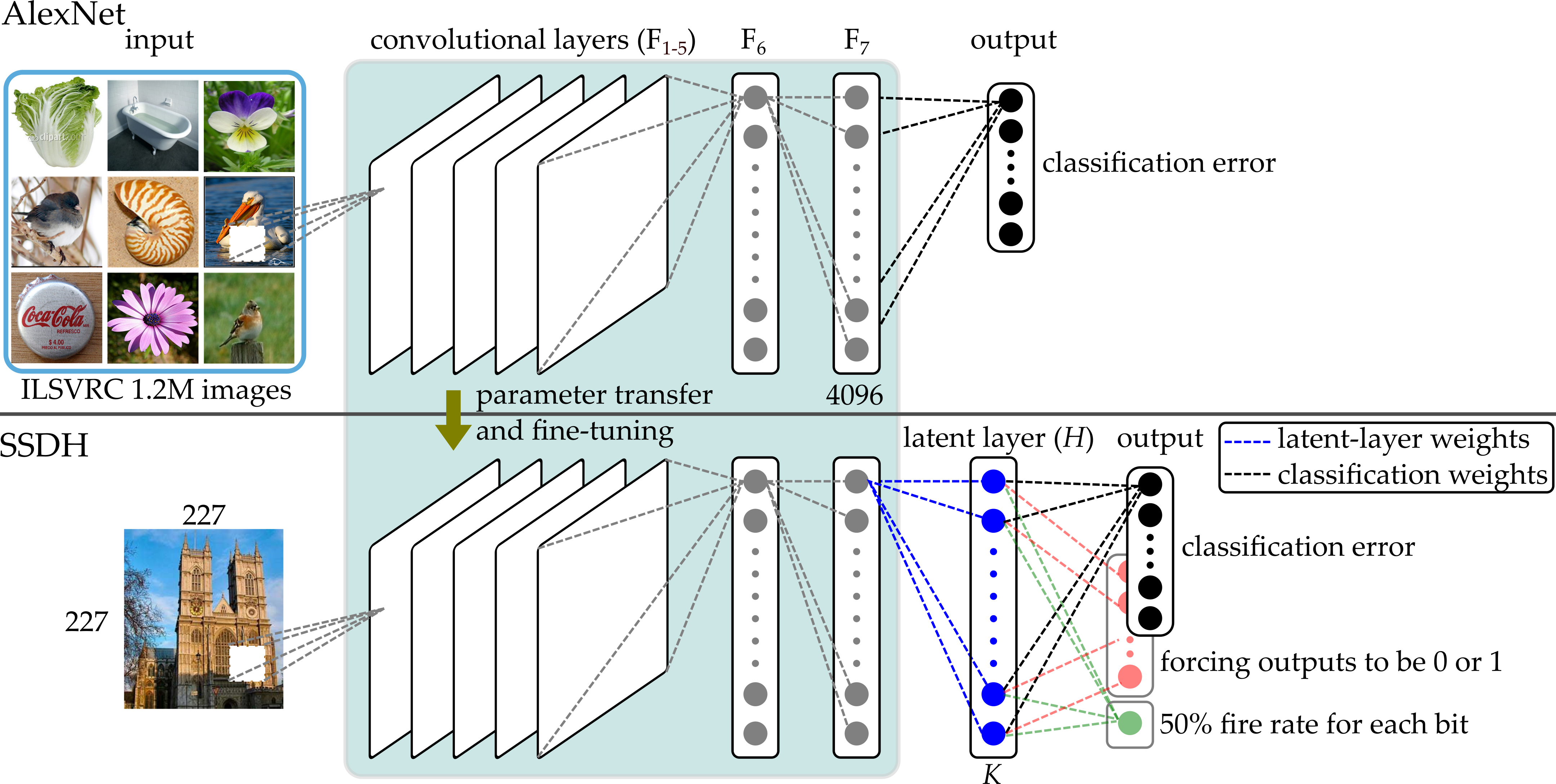}
	\caption{An overview of our proposed supervised semantic-preserving deep hashing (SSDH) \textcolor{black}{that takes AlexNet as an example}.
We construct the hash functions as a latent layer with $K$ units between the image representation layer and classification outputs in a convolutional neural network (CNN). 
SSDH takes inputs from images 
and learns image representations, binary codes, and classification through the optimization of an objective function that combines a classification loss with desirable properties of hash codes.
The learned codes preserve the semantic similarity between images and are compact for image search.}
	\label{fig:deep-hashing}
\end{figure}

In the learning-based hashing approaches, methods based on deep networks~\cite{krizhevsky:esann11,lai:cvpr15,liong:cvpr15,
salakhutdinov:ijar09,xia:aaai14,zhao:cvpr15} form a special group and so we discuss them separately here.
One of the earliest efforts to apply deep networks in hash is semantic hashing (SH)~\cite{salakhutdinov:ijar09}.
It constructs hash codes from unlabeled images via a network with stacked Restricted Boltzmann Machines (RBMs).
The learned binary codes are treated as memory addresses, and thus similar items to a query can be found by simply accessing to memory addresses that are within a Hamming ball around the query vector.
Autoencoders, which aim to learn compressed representations of data, can be used to map images to binary codes.
The deep autoencoder developed in~\cite{krizhevsky:esann11} is initialized with the weights from pre-trained stacks of RBMs, and the code layer uses logistic units whose outputs then are rounded to $1$ or $0$ for binary codes.

Deep networks are also used in deep hashing (DH) and supervised DH (SDH)~\cite{liong:cvpr15} for learning compact binary codes through seeking multiple non-linear projections to map samples into binary codes.
Deep multi-view hashing (DMVH)~\cite{kang:icdm12} constructs a network with view-specific and shared hidden units to handle multi-view data.
\textcolor{black}{However, these methods rely on hand-crafted features, which need strong prior to design beforehand and do not evolve along the code learning.
Our SSDH, by contrast, couples feature learning and code construction in a single model.
Under the semantics supervision, both of them evolve into a feature space where semantically similar contents tend to share similar codes.}
Recently, hashing methods based on CNNs have also been proposed.
CNNH and CNNH+~\cite{xia:aaai14} employ a two-stage learning approach that first decomposes a pairwise similarity matrix into approximate hash codes based on data labels and then trains a CNN for learning the hash functions.
\textcolor{black}{The method in~\cite{lai:cvpr15} and deep semantic ranking based hashing (DSRH)~\cite{zhao:cvpr15} adopt a triplet ranking loss derived from labels for code construction.
Like these approaches, our method also exploits label information in code learning.
However, ours differs from them 
in several ways.
First, our SSDH imposes additional constraints on the latent layer to learn more separated codes while no such constraints are applied in~\cite{lai:cvpr15,zhao:cvpr15}.
Second, ours can be achieved by a slight modification to an existing network while~\cite{lai:cvpr15} requires a more complex network configuration with significant modifications.
Finally, our approach learns in a point-wised manner but} some of these approaches need to perform a matrix factorization prior to hash function learning (\eg, CNNH and CNNH+~\cite{xia:aaai14}) and some need to take inputs in the form of image pairs (\eg, SDH~\cite{liong:cvpr15}) or image triples (\eg,~\cite{lai:cvpr15} and DSRH~\cite{zhao:cvpr15}), which make them less favorable when the data size is large.

\subsection{Supervised Deep Transfer Learning}
In deep learning, the networks can be pre-trained in an unsupervised way based on an energy-based probability model in RBM and deep belief networks~\cite{hinto:neco06}, or via self-reproducing in autoencoders \cite{krizhevsky:esann11}.
Then, followed by supervised training (\ie, fine-tuning) the network can be optimized for a particular task.

Pre-training has been pushed forward to supervised learning recently.
Supervised pre-training and fine-tuning has been employed in CNN and shown promising performance.
It follows the \emph{inductive transfer learning} principle~\cite{pan:tkde10}, which adopts the idea that one cannot learn how to walk before crawl, or how to run before walk.
Hence, the connection strengths trained from one or more tasks for a neural network can be used as initial conditions and further adapted to suit new and/or higher-level tasks in other domains.
Supervised pre-training investigated in DeCAF~\cite{donahue:icml14} shows that a deep CNN pre-trained with supervision on the ImageNet dataset~\cite{deng:cvpr09} can be used as a feature extractor.
The obtained deep convolutional features are effective for other visual tasks, such as scene classification, domain adaptation, and fine-grained recognition.
The capacity of deep representations is investigated in~\cite{oquab:cvpr14}, in which mid-level representations of a pre-trained CNN are transferred and two adaptation layers are added to the top of deep features for learning a new task.
The work shows that transfer learning can be achieved with only limited amount of training data.
Unlike~\cite{oquab:cvpr14} where the fine-tune is only performed in the additional layers for classification, the Region-based Convolutional Network (R-CNN)~\cite{girshick:cvpr14,girshick:pami16} fine-tunes the entire network for domain-specific tasks of object detection and segmentation.

\textcolor{black}{Besides, such deep features have recently gained much attention in image retrieval as well. 
As shown in Krizhevsky \etal~\cite{krizhevsky:nips12}, the features of CNNs learned on large data can be used for retrieval.
Since then, deep features have been widely adopted in image search. 
For example, the work in~\cite{babenko:eccv14} has extensively evaluated the performance of deep features as a global descriptor.
Gong \etal~\cite{gong:eccv14} propose to use Vector of Locally Aggregated Descriptors (VLAD) to pool deep features of local patches at multiple scales.
Babenko and Lempitsky~\cite{babenko:iccv15} suggest a sum-pooling aggregation method to generate compact global descriptors from local deep features, and the work in~\cite{razavian:cvprws14} studies the spatial search strategy to improve retrieval performance.}

How to exploit the strength of supervised deep transfer learning for hash function construction has not been explored yet.
In this paper, instead of performing inductive transfer learning merely for the purpose of task domain conversions, we further investigate the adaptation problem in the functionality level.
The proposed approach fine-tunes the weights to a new domain for classification and also realizes a function-level tuning to generate semantic-aware binary codes.
Our approach relies on an enhancement of existing classification architectures, and we show that the classification performance will not be degraded experimentally.
It thus provides a multi-purpose architecture effective for both retrieval and classification.

\vspace*{-5pt}
\section{Learning Hash Codes via Deep Networks}\label{sec:method}

Let $\mathcal{I} = \left\lbrace I_n \right\rbrace _{n=1}^{N}$ be $N$ images and $\mathcal{Y} = \left\lbrace y_n \in \{0,1\}^M \right\rbrace^{N} $ be their associated label vectors, where $M$ denotes the total number of class labels.
An entry of the vector $y_n$ is 1 if an image $I_n$ belongs to the corresponding class and $0$ otherwise.
Our goal is to learn a mapping
$\mathcal{F}: \mathcal{I} \rightarrow \left\lbrace 0,1 \right\rbrace ^{K \times N}$, which maps images to their $K$-bits binary codes $B = \left\lbrace b_n\right\rbrace  \in \left\lbrace 0,1\right\rbrace^{K \times N}$ while preserving the semantic similarity between image data.
Specifically, we aim to design a supervised hashing algorithm that exploits the semantic labels to 
create binary codes of the following properties:
\begin{itemize}[leftmargin=*]
\item The codes respect the semantic similarity between image labels.
Images that share common class labels are mapped to same (or close) binary codes.
\item The bits in a code are evenly distributed and discriminative. 
\end{itemize}
\vspace*{-5pt}
\subsection{Deep Hashing Functions}\label{subsec:deep-hashing}
We take advantage of recent advances in deep learning and construct the hash functions on a CNN that is capable of learning semantic representations from images.
\textcolor{black}{
Our approach is based on existing deep models, such as AlexNet~\cite{krizhevsky:nips12} and VGG~\cite{simonyan:iclr15}.
It can be integrated with other deep models as well.
Without loss of generality, we introduce our approach based on AlexNet in the following.}

The architecture of AlexNet is illustrated in the top half of Figure~\ref{fig:deep-hashing}.
It has $5$ convolution layers ($F_{1-5}$) with max-pooling operations followed by $2$ fully connected layers ($F_{6-7}$) and an output layer.
In the convolutional layers, units are organized into feature maps and are connected locally to patches in the outputs (\ie, feature maps) of the previous layer.
The fully-connected layers can be viewed as a classifier when the task is to recognize images.
The convolution and first two fully-connected layers ($F_{6-7}$) are composed of the rectified linear units (ReLUs) because the ReLUs lead to faster training. 
AlexNet is designed in particular for multi-class classification problems so that its output layer is a classification layer have the units of the same number of class labels. 
The output units are with the softmax functions and the network is trained to maximize the multinomial logistic regression objective function for multi-class classification.
To incorporate the deep representations into the hash function learning, we add a \emph{latent layer} $H$ with $K$ units to the top of layer $F_7$ \textcolor{black}{(i.e., the layer right before the output layer)}, as illustrated in the bottom half of Figure~\ref{fig:deep-hashing}.
This latent layer is fully connected to $F_7$ and uses the sigmoid units so that the activations are between $0$ and $1$.

Let $W^{H} \in \mathbb{R}^{d \times K}$ denote the weights (\ie the projection matrix) between $F_7$ and the latent layer.
For a given image $I_n$ with the feature vector $a_n^{7} \in \mathbb{R}^{d}$ in layer $F_7$, the activations of the units in $H$ can be computed as $a_n^{H} = \sigma(a_n^{7}W^H + b^{H})$, where $a_n^{H}$ is a $K$-dimensional vector, $b^{H}$ is the bias term and $\sigma(\cdot)$ is the logistic sigmoid function, defined by $\sigma(z) = 1/(1+\mathrm{exp}(-z))$, with $z$ a real value.
The binary encoding function is given by
\begin{align}\label{eqn:binarization}
	b_n & = (\mathrm{sgn}(\sigma(a_n^{7}W^H + b^{H}) - 0.5)+1)/2 \nonumber \\
        & = (\mathrm{sgn}(a_n^{H} - 0.5)+1)/2,
\end{align}
where $\mathrm{sgn}(v) = 1$ if $v > 0$ and $-1$ otherwise, and $\mathrm{sgn}(\cdot)$ performs element-wise operations for a matrix or a vector.

\subsection{Label Consistent Binary Codes}\label{subsec:label-consistency}
Image labels not only provide knowledge in classifying images but also are useful supervised information for learning hash functions.
We propose to model the relationship between the labels and the binary codes in order to construct semantics-preserving binary codes.
We assume that the semantic labels can be derived from a set of $K$ latent concepts (or hidden attributes) with each attribute \emph{on} or \emph{off}.
When an input image is associated with binary-valued outputs (in $\{0,1\}^K$), the classification is dependent on these hidden attributes.
This implies that through an optimization of a loss function defined on the classification error, we can ensure that semantically similar images are mapped to similar binary codes.

Consider a matrix $W^C \in \mathbb{R}^{K \times M}$ that performs a linear mapping of the binary hidden attributes to the class labels.
Incorporating such a matrix into our the network amounts to adding a classification layer to the top of the latent layer (see Figure~\ref{fig:deep-hashing} where the black dashed lines denote $W^{C}$).
Let $\hat{y}_n$ denote the prediction of our network (the black nodes in Figure~\ref{fig:deep-hashing}) for an image $I_n$.
In terms of the classification formulation, to solve $W^C$, one can choose to optimize the following objective function:
\begin{equation}\label{eqn:classification}
	\argmin_{W} E_1(W)=
\argmin_{W} \sum_{n=1}^{N} L(y_n, \hat{y}_{n}) + \lambda ||W||^2,
\end{equation}
where $L(\cdot)$ is a loss function that minimizes classification error and will be detailed below, $W$ denotes the weights of the network, and $\lambda$ governs the relative importance of the regularization term.

The choice of the loss function depends on the problem itself.
For multi-class classification, we simply follow the setting in AlexNet that uses softmax outputs and minimizes the cross-entropy error function:
\begin{equation}
	L(y_n, \hat{y}_{n}) = -\sum_{m=1}^{M} y_{nm} \mathrm{ln} \hat{y}_{nm},
\end{equation}
where $y_{nm}$ and $\hat{y}_{nm}$ are the desired output and the prediction of the $m$th unit, respectively.

We introduce a maximum-margin loss function to fulfill the goal of multi-label classification because the loss function in AlexNet is designed only for the single-label purpose.
Following the same notions, let $\mathcal{Y} = \left\lbrace y_{nm} \right\rbrace^{N \times M}$ denote the label vectors associated with $N$ images of $M$ class labels.
In multi-label classification, an image is associated with multiple classes and thus multiple entries of $y_n$ could be $1$, and the outputs in our network are $m=\{1, \cdots M\}$ binary classifiers.
Given the $n$-th image sample with the label $y_{nm}$, we want the $m$-th output node of the network to have positive response for the desired label $y_{nm}=1$ (\ie., positive sample) and negative response for $y_{nm}=0$ (\ie., negative sample).
In specific, to enlarge the margin of the classification boundary, for samples of a particular label $y_{nm}$, we set the network to have the outputs $\hat{y}_{nm} \geq 1$ for $y_{nm} = 1$ and $\hat{y}_{nm} \leq 0$ for $y_{nm} = 0$.
The loss $l(y_{nm}, \hat{y}_{nm})$ for each output node is defined as
\begin{equation}\label{eqn:multi-label_loss}
	l(y_{nm}, \hat{y}_{nm}) = \\
	\begin{cases}
	0                          & y_{nm}=1 \wedge \hat{y}_{nm} \geq 1\\
    0                          & y_{nm} = 0 \wedge \hat{y}_{nm} \leq 0\\
	\textcolor{black}{\frac{1}{2}|y_{nm} - \hat{y}_{nm}|_p^p}  & \text{otherwise}
	\end{cases},
\end{equation}
\textcolor{black}{where $p\in\{1,2\}$.}
\textcolor{black}{
When $p=1$ (or 2), such a loss function actually implements linear L1-norm (or L2-norm) support vector machine (SVM)~\cite{hsieh:icml08} 
thresholded at 0.5.
}
Hence, our network combines the AlexNet architecture, binary latent layer, and SVM classifiers in a cascade for multi-label classification.
Note that to train a large scale linear SVM, the state-of-the-art methods~\cite{hsieh:icml08,fan:jmlr08} employ the coordinate-descent optimization in the dual domain (DCD) of SVM, which is proven to be equivalent to performing stochastic gradient descent (SGD) in the primal domain~\cite{hsieh:icml08}.
As SGD is a standard procedure for training neural networks, when our network is trained only for the SVM layer and the parameters of the other layers are fixed, it is equivalent to solving the convex quadratic programming problem of SVM by using the primal domain SGD method in~\cite{hsieh:icml08,fan:jmlr08} (with SGD's learning rate corresponding to some SVM's model parameter $C$).
When training the entire network, the parameters then evolve to more favorable feature representations (in the AlexNet architecture), latent binary representations (in the hidden layer), and binary classifiers (in the SVMs layer) simultaneously.
The 
gradient with 
the activation of output unit $m$, $\frac{\partial l(y_{nm}, \hat{y}_{nm})}{\partial{\hat{y}_{nm}}}$, takes the form
\begin{equation}
	\delta_{m} = 
	\begin{cases}
	0                          \hfill y_{nm} = 1 \wedge \hat{y}_{nm} \geq 1 \\
	0                          \hfill y_{nm} = 0 \wedge \hat{y}_{nm} \leq 0 \\
	\textcolor{black}{\frac{p}{2}\mathrm{sgn}(\hat{y}_{nm} - y_{nm}) |\hat{y}_{nm} - y_{nm}|^{p-1}}  \quad \text{otherwise}
	\end{cases},
\end{equation}
\textcolor{black}{for $p=1$ or 2}. 
Because the loss function 
is \textcolor{black}{almost} differentiable everywhere, it is  suitable for gradient-based optimization methods.
Finally, the loss function $L(y_n, \hat{y}_{n})$ is defined as the summation of the losses of output units,
\begin{equation}
	L(y_n, \hat{y}_{n}) = \sum_{m=1}^{M}l(y_{nm}, \hat{y}_{nm}).
\end{equation}

\subsection{Efficient Binary Codes}\label{subsec:efficient-binary-codes}
Apart from that semantically similar images have similar binary codes, we encourage the activation of each latent node to approximate to $\left\lbrace 0,1 \right\rbrace$.
Let $a_{nk}^H$ ($k=1, \cdots, K$) be the $k$-th element of the hidden vector $a_n^H$.
Because $a_{nk}^{H}$ has already been activated by a sigmoid function, its value is inside the range [0,1].
To further make the codes approach to either $0$ or $1$, it can be achieved by adding the constraint of maximizing the sum of squared errors between the latent-layer activations and $0.5$, that is, $\sum_{n=1}^{N} ||a_n^{H} - 0.5 \mathbf{e}||^{2}$, where $\mathbf{e}$ is the $K$-dimensional vector with all elements $1$.
With this constraint, the codes generated by our network can fulfill the binary-valued requirement more appropriately.

Besides making the codes binarized, we consider further the balance property.
This could be achieved by letting $50\%$ of the values in the training samples $\{a_{nk}^H\}_{n=1}^N$ be $0$ and the other $50\%$ be $1$ for each bit $k$ as suggested in \cite{weiss:nips08}.
However, because all of the training data are jointly involved to fulfill this constraint, it is difficult to be implemented in mini-batches when SGD is applied for the optimization. 

In this paper, we want to keep the constraints decomposable to sample-wised terms so that they are realizable with SGD in a point-wised way.
To make the binary codes balanced, we consider a different constraint implementable with mini-batches.
Given an image $I_n$, let $\{a_{nk}^H\}_{k=1}^K$ form a discrete probability distribution over \{0, 1\}.
We hope that there is no preference for the hidden values to be $0$ or $1$.
That is, the occurrence probability of each bit's on or off is the same, or the entropy of the discrete distribution is maximized.
To this end, we want each bit to fire $50\%$ of the time via minimizing $\sum_{n=1}^{N}(\mathrm{mean}(a_n^{H}) - 0.5)^2$, where $\mathrm{mean}(\cdot)$ computes the average of the elements in a vector.
The criterion thus favors binary codes with an equal number of $0$'s and $1$'s in the learning objective.
It also enlarges the minimal gap and makes the codes more separated because the minimal Hamming distance between two binary strings with the same amounts of $0$'s and $1$'s is $2$ (but not $1$).

In sum, combining these two constraints makes $a_{n}^{H}$ close to a length-$K$ binary string with a $50\%$ chance of each bit being $0$ or $1$, and we aim to optimize the following objective to obtain the binary codes:
\begin{align}\label{eqn:efficient-codes}
	&\argmin_{W}  -\frac{1}{K}\sum_{n=1}^{N} ||a_n^{H} - 0.5 \mathbf{e} ||\textcolor{black}{_p^p} + \sum_{n=1}^{N}|\mathrm{mean}(a_n^{H}) - 0.5|\textcolor{black}{^p} \nonumber \\
     = &\argmin_{W} -E_2(W)+E_3(W),
\end{align}
\textcolor{black}{where $p\in\{1,2\}$}.
The first term encourages
the activations of the units in $H$ to be close to either $0$ or $1$, and the second term further ensures that the output of each node has a nearly $50\%$ chance of being $0$ or $1$.
Note that the objective designed in Eq.~\eqref{eqn:efficient-codes} remains a sum-of-losses form.
It keeps the property that each loss term is contributed by only an individual training sample and no cross-sample terms are involved in the loss function.
Hence, the objective remains point-wised and can be minimized through SGD efficiently by dividing the training samples (but not pairs or triples of them) into batches. 
Our network thus relies on the minimization of a latent-concept-driven classification objective with some sufficient conditions on the latent codes to learn semantic-aware binary representations, which can be shown fairly effective on various datasets in our experiments.

On the network design, we add a unit (the green node in the bottom half of Figure~\ref{fig:deep-hashing}) that performs an average pooling operation (the green dashed lines) over the nodes in the latent layer to obtain the mean activation for 
the $E_3(\cdot)$ term in Eq.~\eqref{eqn:efficient-codes}.
The weights associated with the connections to this unit are fixed to $1/K$.
The $E_2(\cdot)$ term in Eq.~\eqref{eqn:efficient-codes} imposes constraints directly on the units in the latent layer.
No modification to the network is needed.
However, for the clarity of presentation, we draw additional red nodes in Figure~\ref{fig:deep-hashing} to indicate this constraint.
\subsection{Overall Objective and Implementation}
The entire objective function aiming for constructing similarity preserving ($E_1(W)$ in Eq.~\eqref{eqn:classification}) and binarization properties (Eq.~\eqref{eqn:efficient-codes}) is given as:
\begin{equation}\label{eqn:overall-objective}
	\argmin_{W} \quad \alpha E_1(W) - \beta E_2(W) + \gamma E_3(W),
\end{equation}
where $\alpha$, $\beta$, and $\gamma$ are 
the weights of each term.

We implement our approach by using the open source CAFFE~\cite{jia:acmmm14} package with an 
\textcolor{black}{NVIDIA Titan X} GPU.
{\color{black}
To optimize \eqref{eqn:overall-objective}, in addition to the output layer for classification, we add two new loss layers for 
$E_2$ and $E_3$, respectively, 
on top of the latent layer.
When performing multi-label classification, 
the output layer is replaced with the maximum-margin loss layer in our implementation.}
As our network is adapted from AlexNet~\cite{krizhevsky:nips12} that has been trained on the $1.2$ million ILSVRC subset of the ImageNet for the $1000$-class recognition task, the initial weights in layers $F_{1-7}$ of our network are set as the pre-trained ones and the remaining weights are randomly initialized.
We apply SGD, in conjunction with backpropagation, with mini-batches to network training for minimizing the overall objective in Eq.~\eqref{eqn:overall-objective}.
We also employ dropout in which the activations of the intermediate units are set to zero with a probability of $0.5$ during training in order to avoid over-fitting.
\textcolor{black}{The parameters $\alpha$, $\beta$, and $\gamma$ 
are evaluated on a dataset at first, and then all are set as $1$ in our experiments.}
Our model is a lightweight modification of \textcolor{black}{an existing network} and thus is easy to implement.
The codes are publicly available\footnote{https://github.com/kevinlin311tw/Caffe-DeepBinaryCode}.

\noindent \textbf{Relation to ``AlexNet feature + LSH"}: The relationship between our approach and an naive combination, \emph{AlexNet feature + LSH} is worth a mention.
Because random Gaussian weights are used for initializing the weights between $F_7$ and the latent layer,  our network can be regarded as initialized with LSH (i.e., random weights) to map the deep features learned in ImageNet (AlexNet feature) to binary codes.
Through SGD learning, the weights of the pre-trained, latent, and classification layers evolve a multi-layer function more suitable for the new domain.
Compared to the straightforward combination of AlexNet features and LSH, our approach can obtain more favorable results as demonstrated in the experiments in Section~\ref{sec:exp}.
\subsection{Binary Codes for Retrieval}

\begin{figure}[tb]
	\centering
	\includegraphics[width=0.75\columnwidth]{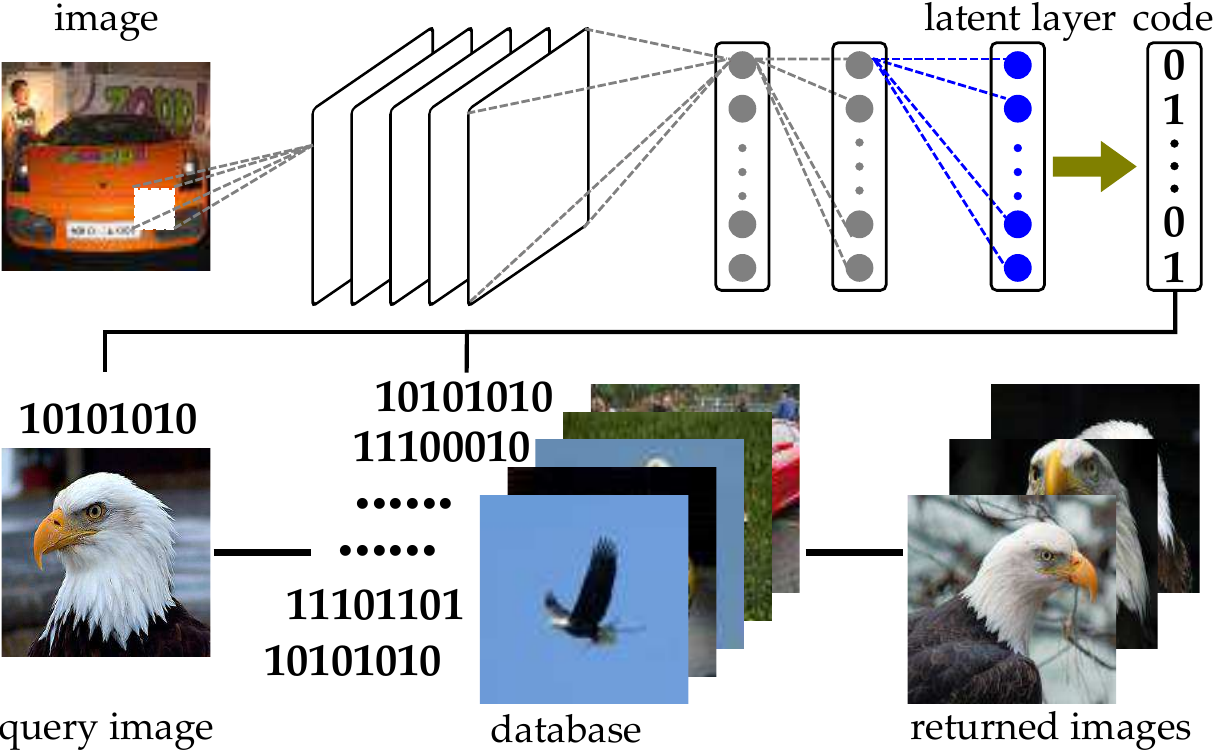}
	\caption{Binary codes for retrieval. Images are fed to the network, and their corresponding binary codes are obtained by binarizing the activations of the latent layer. For image retrieval, the binary codes of a query and of every image in the database are compared based on the Hamming distance. The images 
closest to the query are returned as the results.}
	\label{fig:deep-hashing-retrieval}
\end{figure}

Figure~\ref{fig:deep-hashing-retrieval} illustrates the scheme used to extract binary codes and retrieve similar images for a query.
First, images are fed to the network, and the activations of the latent layer are extracted.
Then, the binary codes are obtained by quantizing the extracted activations via Eq.~\eqref{eqn:binarization}.
Similar images to a novel query are found by computing the Hamming distances between the binary codes of the query and the database images and selecting the images with small Hamming distances in the database as retrieval results. 
\vspace*{-5pt}
\section{Experiments}\label{sec:exp}
We conduct experiments on several benchmarks to compare our method with the state-of-the-art methods.
We also apply our method to \textcolor{black}{large datasets} containing more than 1 million images to show its scalability.
The images in the 
datasets are in a wide spectrum of image types including tiny objects of CIFAR-10, web images of NUS-WIDE, handwritten digits of MNIST, catalog images of UT-ZAP50K, as well as scene images of SUN397, \textcolor{black}{Oxford, and Paris}.
The large datasets, Yahoo-1M and \textcolor{black}{ILSVRC, comprise product and object images with heterogeneous types, respectively.}
The evaluation protocols and datasets are summarized as follows.

\subsection{Evaluation Protocols}
We use three evaluation metrics widely adopted in the literature for the performance comparison.
They measure the performance of hashing algorithms from different aspects.
\begin{itemize}[leftmargin=*]
\item Mean average precision (mAP): We rank all the images according to their Hamming distances to the query and compute the mAP.
The mAP computes the area under the recall-precision curve and is an indicator of the overall performance of hash functions;
\item Precision at $k$ samples: It is computed as the percentage of true neighbors among the top $k$ retrieved images;
\item Precision within Hamming radius $r$: We compute the precision of the images in the buckets that fall within the Hamming radius $r$ of the query image, where $r=2$ is selected as previous works did.
\end{itemize}
\textcolor{black}{Following the common settings of evaluating the performance of hash methods, we use the class labels as the ground truth and all the above three metrics are computed through examining whether the returned images and the query share a common class label.
For the datasets lacking of class labels, the performance is evaluated via the ground-truth retrieval lists provided for the queries in their test sets.}
\subsection{Datasets}
\begin{figure}[tb]
	\centering
	\renewcommand{\arraystretch}{0.5}
	\setlength{\tabcolsep}{0pt}
	\begin{tabular*}{.95\columnwidth}{@{\extracolsep{\fill}}*{7}c@{}}  
	\includegraphics[width=.1\columnwidth,height=.1\columnwidth]{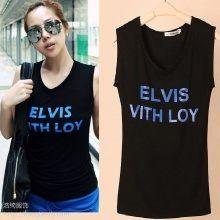}\includegraphics[width=.1\columnwidth,height=.1\columnwidth]{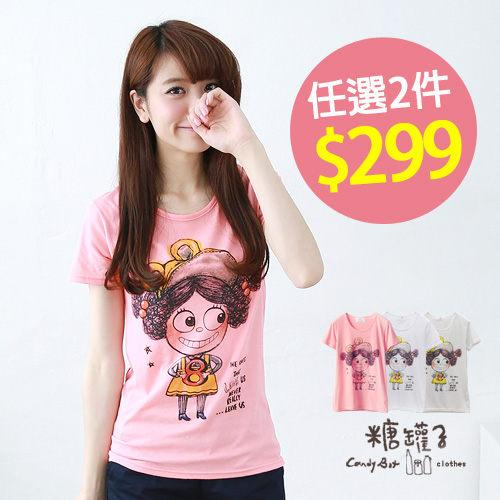}&&
	\includegraphics[width=.1\columnwidth,height=.1\columnwidth]{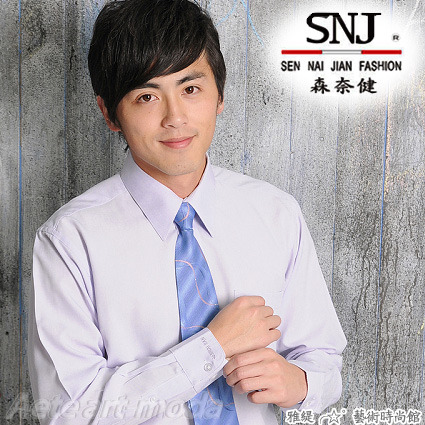}\includegraphics[width=.1\columnwidth,height=.1\columnwidth]{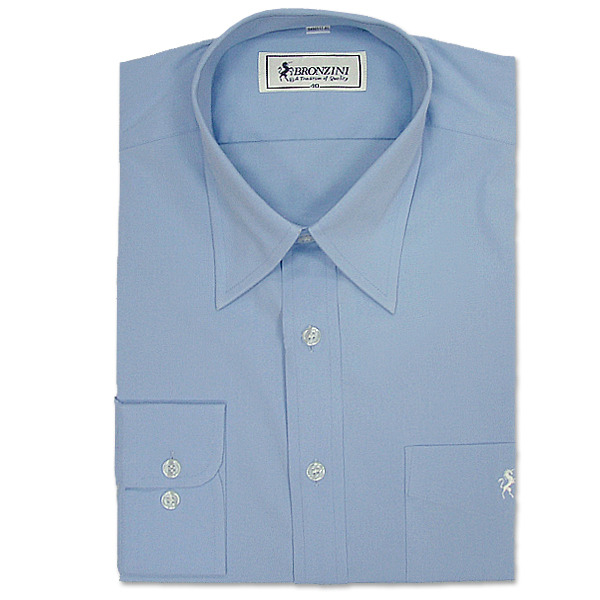}&&
	\includegraphics[width=.1\columnwidth,height=.1\columnwidth]{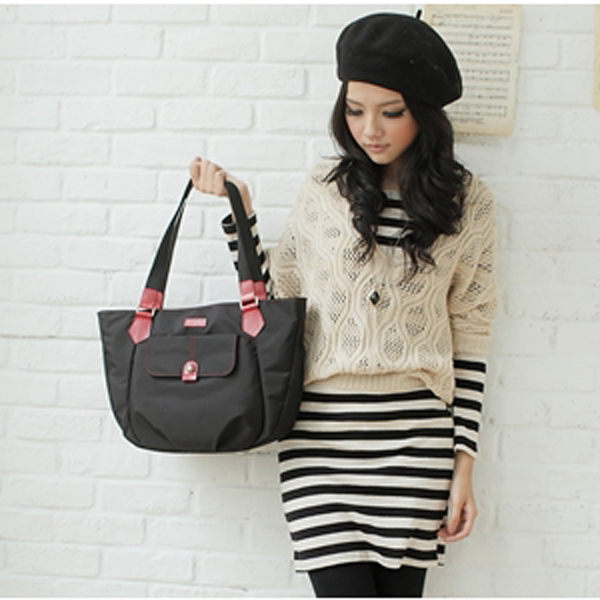}\includegraphics[width=.1\columnwidth,height=.1\columnwidth]{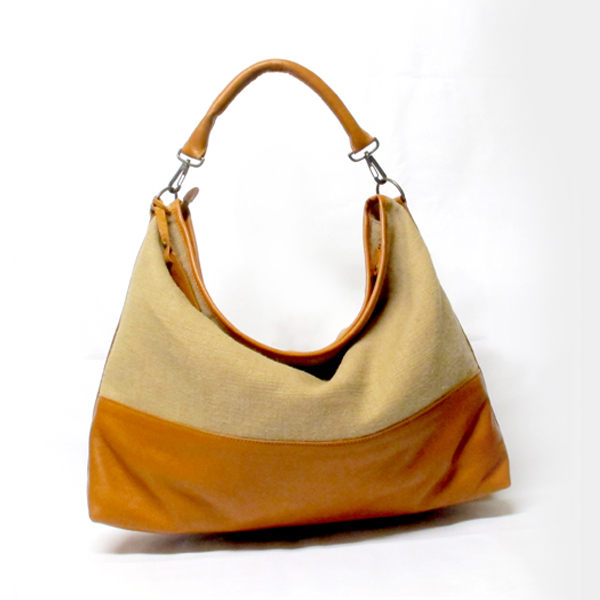} &&
	\includegraphics[width=.1\columnwidth,height=.1\columnwidth]{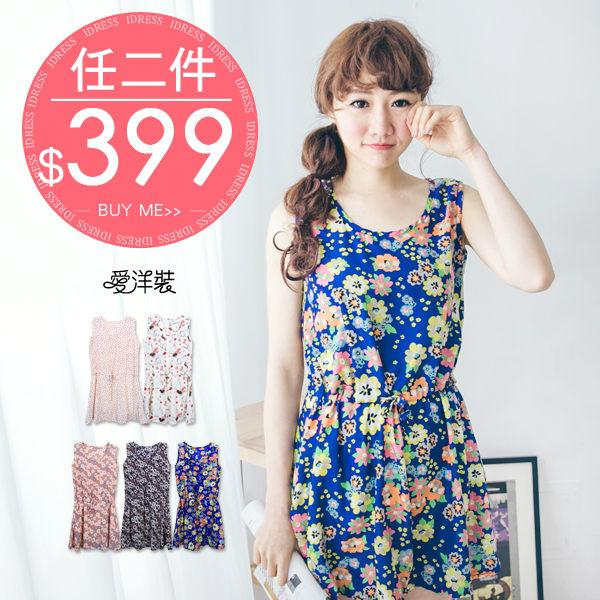}\includegraphics[width=.1\columnwidth,height=.1\columnwidth]{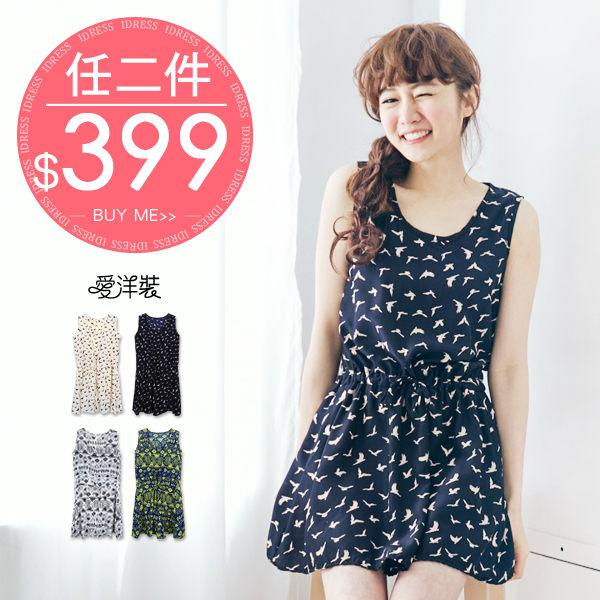}\\
	\footnotesize{top} & & \footnotesize{shirt} & & \footnotesize{bag} & & \footnotesize{dress}\\[5pt]
	\includegraphics[width=.1\columnwidth,height=.1\columnwidth]{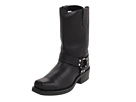}\includegraphics[width=.1\columnwidth,height=.1\columnwidth]{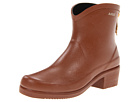}&&
	\includegraphics[width=.1\columnwidth,height=.1\columnwidth]{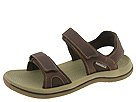}\includegraphics[width=.1\columnwidth,height=.1\columnwidth]{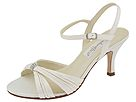}&&
	\includegraphics[width=.1\columnwidth,height=.1\columnwidth]{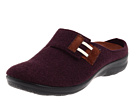}\includegraphics[width=.1\columnwidth,height=.1\columnwidth]{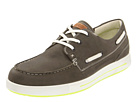}&&
	\includegraphics[width=.1\columnwidth,height=.1\columnwidth]{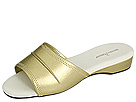}\includegraphics[width=.1\columnwidth,height=.1\columnwidth]{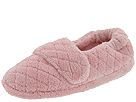}\\
	\footnotesize{boots} & & \footnotesize{sandals} & & \footnotesize{shoes} &  & \footnotesize{slippers}\\
	\end{tabular*}
	\caption{Sample images from the Yahoo-1M and UT-ZAP50K datasets. Upper: Yahoo-1M images. The product images are of heterogeneous types, including those that are backgroundless or of cluttered backgrounds, with or without humans. Lower: UT-ZAP50K images.}
	\label{fig:sample-images-datasets}
\end{figure}
\begin{table}[t]
	\centering
	\caption{Statistics of datasets used in the experiments.}
	\label{tbl:datasets}
	\setlength{\tabcolsep}{0pt}
	\begin{tabular*}{\columnwidth}{@{}@{\extracolsep{\fill}}llrrr@{}}
	\toprule
	Dataset  & Label Type   & \# Labels & Training & Test\\
	\midrule
	CIFAR-10 & Single label & 10       & 50,000           & 1,000\\
	NUS-WIDE & Multi-label  & 21       & 97,214           & 65,075\\
	MNIST    & Single label & 10       & 60,000           & 10,000\\
	SUN397   & Single label & 397      & 100,754          & 8,000\\
	UT-ZAP50K& Multi-label  & 8        & 42,025           & 8,000\\
	Yahoo-1M & Single label & 116      & 1,011,723        & 112,363\\
	\textcolor{black}{ILSVRC2012}&\textcolor{black}{Single label} & \textcolor{black}{1,000}    & \textcolor{black}{$\sim$1.2 M}        & \textcolor{black}{50,000}\\
	\textcolor{black}{Paris} & \textcolor{black}{unsupervised} & \textcolor{black}{N/A}& \textcolor{black}{N/A} & \textcolor{black}{55}\\
	\textcolor{black}{Oxford} & \textcolor{black}{unsupervised} & \textcolor{black}{N/A} & \textcolor{black}{N/A} & \textcolor{black}{55}\\
	\bottomrule
	\end{tabular*}
\end{table}

\noindent \textbf{CIFAR-10~\cite{krizhevsky:tr09}} is a dataset consists of 60,000 $32 \times 32$ color images categorized into 10 classes.
The class labels are mutually exclusive, and thus each class has 6,000 images.
The entire dataset is partitioned into two non-overlapping sets: a training set with 50,000 images and a test set with 10,000 images.
Following the settings in~\cite{lai:cvpr15,xia:aaai14}, we randomly sampled 1,000 images, 100 images per class, from the test set to form the query set for performance evaluation.
\textcolor{black}{CIFAR-10 is one of the most commonly used datasets for evaluating hash-based image retrieval approaches.}

\noindent \textbf{NUS-WIDE~\cite{chua:civr09}} is a dataset comprising about 270,000 images collected from Flickr.
Each image belongs to more than one category taken from 81 concept tags.
The NUS-WIDE website provides only the URLs of images, and following the given links, we were able to collect about 230,000 images as the other images have been removed by the owners.
Following the settings in~\cite{lai:cvpr15,xia:aaai14}, we use images in the 21 most frequent labels, with at least 5,000 images per label, in the evaluation.
The downloaded images are divided into a training set of 97,214 images and a test set of 65,075 images.
The training set is used for network training, and in accordance with the evaluation protocols used in~\cite{lai:cvpr15,xia:aaai14}, 100 images per label are randomly sampled from the test set to form a query set of 2,100 images.

\noindent \textbf{MNIST} is a dataset of 70,000 $28 \times 28$ grayscale images of handwritten digits grouped into 10 classes.
It comprises 60,000 training and 10,000 testing images.

\noindent \textbf{SUN397~\cite{xiao:cvpr10}
} is a large scene dataset consisting of 108,754 images in 397 categories.
The number of images varies across categories, with each category containing at least 100 images.
Following the settings in~\cite{lin:cvpr14}, we randomly select 8,000 images to form the query set and use the remaining 100,754 as the training samples.

\noindent \textbf{UT-ZAP50K~\cite{yu:cvpr14}
} consists of 50,025 catalog images collected from Zappos.com.
Some selected images are shown in Figure~\ref{fig:sample-images-datasets}.
This dataset is created for fine-grained visual comparisons on a shopping task.
To use it in a retrieval task, we associate images with multiple labels from 8 selected classes (4 categories (boots, sandals, shoes, and slippers) and 4 gender labels (boys, girls, men, and women)).
We randomly select 8,000 images, 1,000 per class, as the test set and use the remaining images (42,025) for training.

\noindent \textbf{Yahoo-1M Shopping Images
} contains 1,124,086 product images of heterogeneous types collected from the Yahoo shopping sites.
The images are of cluttered backgrounds or backgroundless, with or without humans.
Figure~\ref{fig:sample-images-datasets} shows some selected images.
Each image is associated with a class label, and there are 116 classes in total.
The number of images in each class varies greatly, ranging from 1,007 to 150,211.
To divide the dataset into two sets, we selected 90\% of the images from each class as training samples and the rest 10\% as test samples.
The entire dataset is thus partitioned into a training set of 1,011,723 images and a test set of 112,363 images.

\noindent \textcolor{black} {\noindent \textbf{ILSVRC2012~\cite{russakovsky:ijcv15}} is the 
dataset for the ImageNet Large Scale Visual Recognition Challenge, 
and also the dataset used for pre-raining the AlexNet and VGG network models 
available on CAFFE.
It has 1,000 object classes 
with approximately 1.2 million training images, 50,000 validation images, and 100,000 test images.
Each image contains a salient object, and the objects in this dataset tend to be centered in the images.
We use the training set for network learning and employ the validation set as the query in the evaluation.
}

\noindent \textcolor{black} {\textbf{Paris~\cite{philbin:cvpr08}} is a standard benchmark for instance-level image retrieval.
It includes 6,412 images of Paris landmarks.
The performance of retrieval algorithms is measured based on the mAP of 55 queries.
}

\noindent \textcolor{black} {\textbf{Oxford~\cite{philbin:cvpr07}} is another widely used benchmark for instance-level image retrieval.
It consists of 5,062 images corresponding to 11 Oxford landmarks.
Images are with considerable variations in viewpoints and scales, thereby making Oxford a more challenging dataset than Paris.
Like Paris, 55 queries (5 per landmark) are used for performance evaluation.
}

Information of these datasets can be found in Table~\ref{tbl:datasets}.
Note that our network takes fixed-sized image inputs.
Images of all datasets are normalized to $256 \times 256$ and then center-cropped to $227 \times 227$ as inputs \textcolor{black}{to AlexNet and $224 \times 224$ to VGG, respectively, following the associated models that are pre-trained and available on CAFFE.
Unless otherwise mentioned, the results are conducted by using our SSDH on the AlexNet architecture.}

\begin{figure*}[t]
	\centering
	\subfloat{\label{fig:cifar-results-map}\small{(a)}\raisebox{-.95\height}{\includegraphics[height=.205\textwidth]{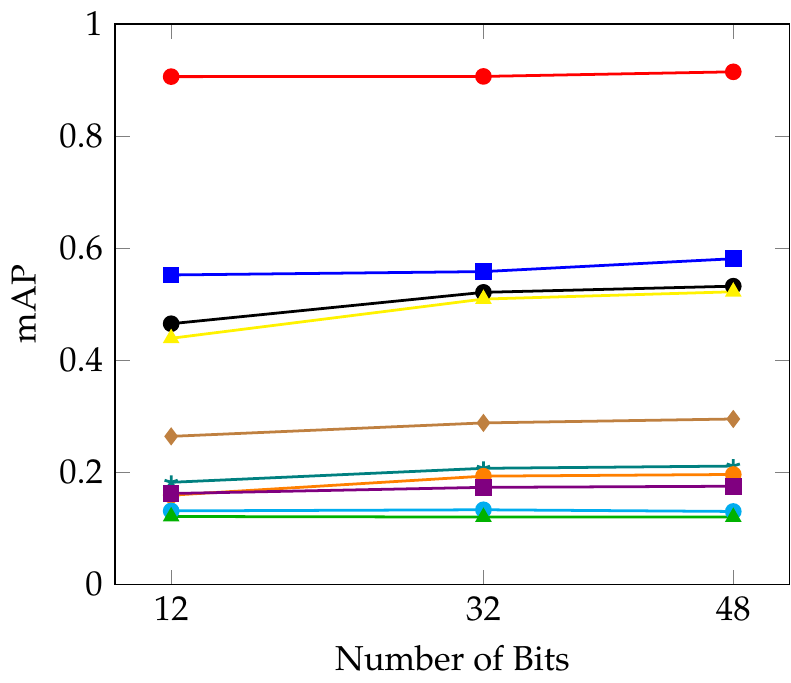}}}\qquad
	\subfloat{\label{fig:cifar-results-precision-k}\small{(b)}\raisebox{-.95\height}{\includegraphics[height=.205\textwidth]{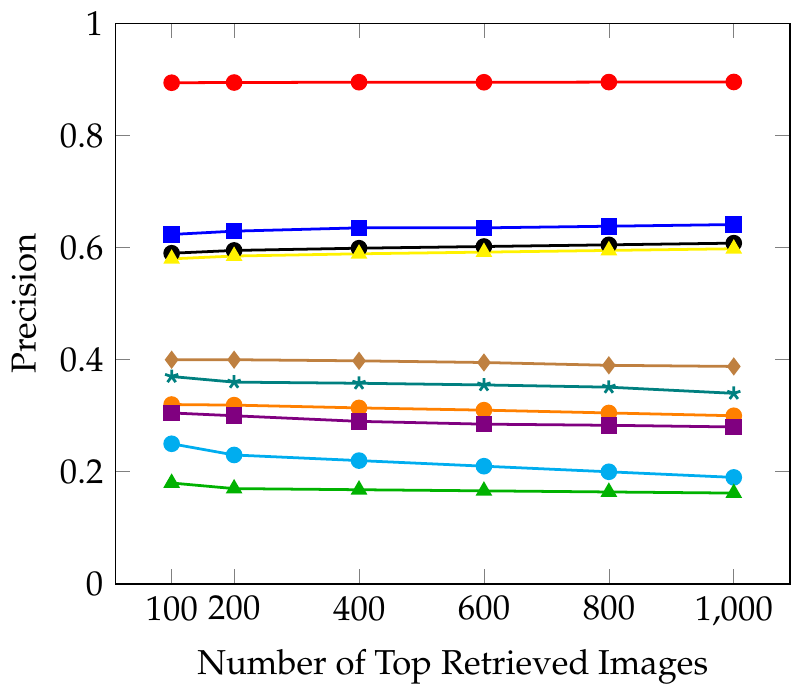}}}\qquad
	\subfloat{\label{fig:cifar-results-hammingradius}\small{(c)}\raisebox{-.95\height}{\includegraphics[height=.205\textwidth]{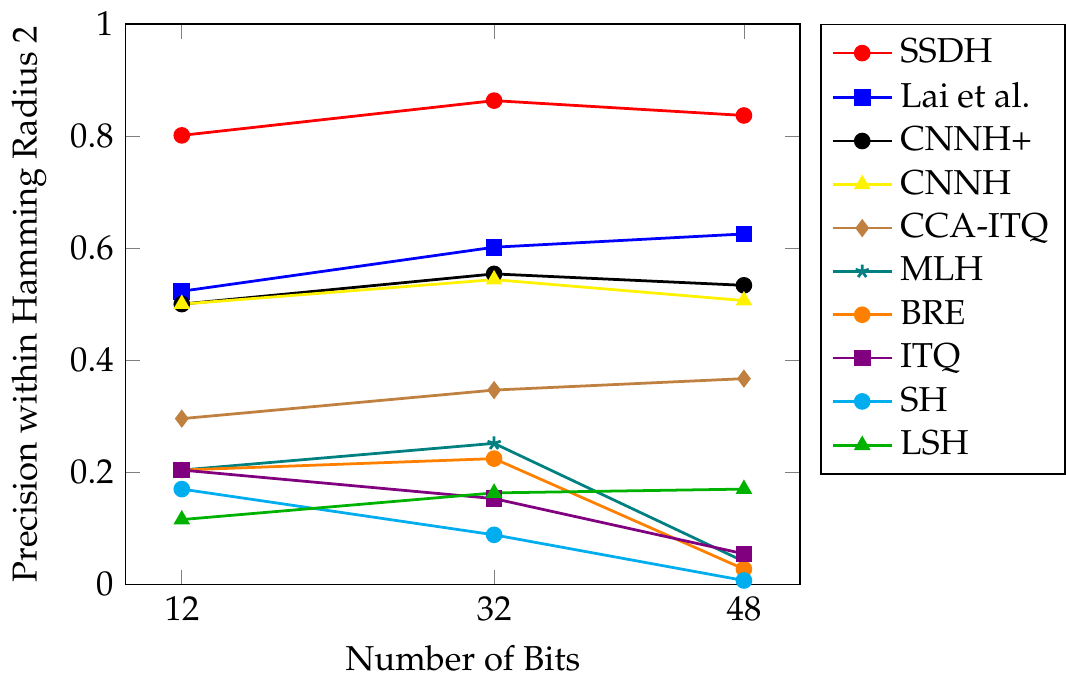}}}
	\caption{Comparative evaluation of different hashing algorithms on the CIFAR-10 dataset. (a) mAP curves with respect to different number of hash bits. (b) Precision curves with respect to different number of top retrieved samples when the 48-bit hash codes are used in the evaluation. (c) Precision within Hamming radius 2 curves with respect to different number of hash bits.}
	\label{fig:cifar-results}
\end{figure*}
\begin{figure*}[tb]
	\centering
	\subfloat{\label{fig:mnist-results-map}\small{(a)}\raisebox{-.95\height}{\includegraphics[height=.198\textwidth]{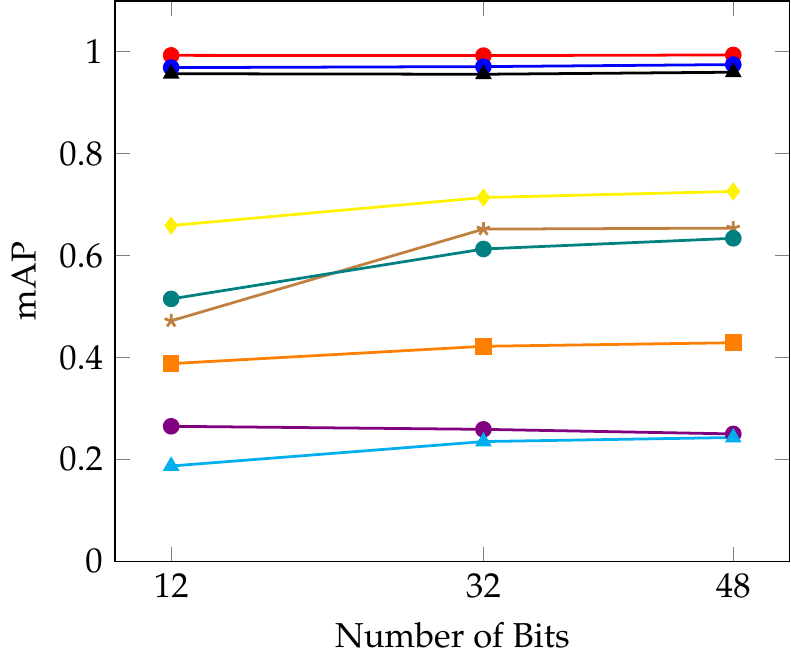}}}\qquad
	\subfloat{\label{fig:mnist-results-precision-k}\small{(b)}\raisebox{-.95\height}{\includegraphics[height=.198\textwidth]{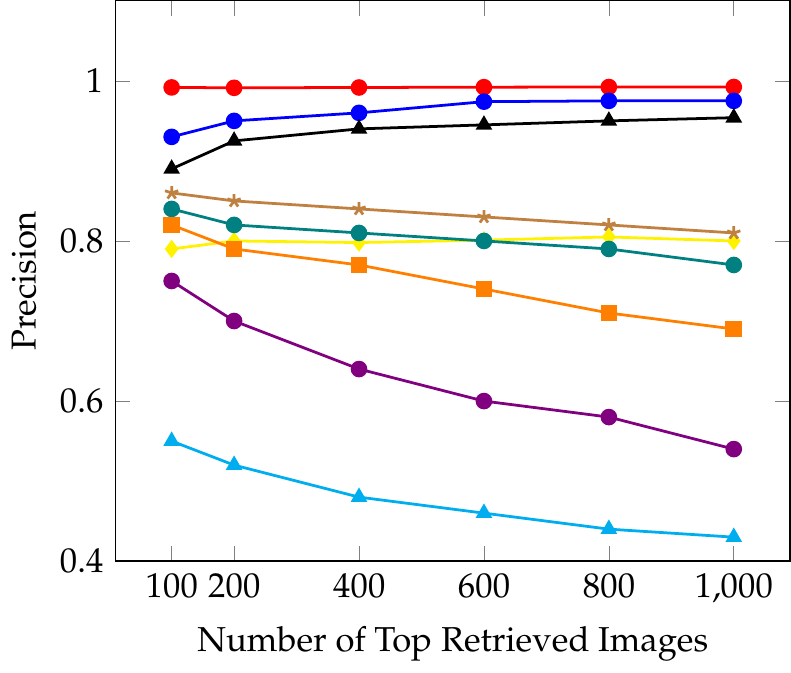}}}\qquad
	\subfloat{\label{fig:mnist-results-hammingradius}\small{(c)}\raisebox{-.95\height}{\includegraphics[height=.198\textwidth]{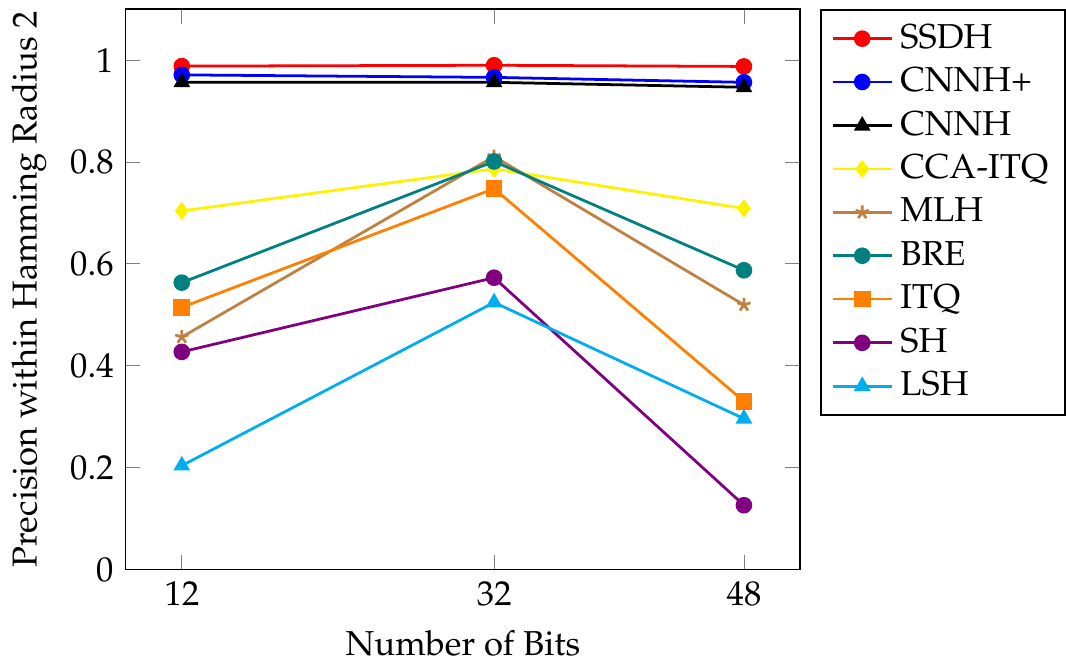}}}
	\caption{Comparative evaluation of different hashing algorithms on the MNIST dataset. (a) mAP curves with respect to different number of hash bits. (b) Precision curves with respect to different number of top retrieved samples when the 48-bit hash codes are used in the evaluation.
(c) Precision within Hamming radius 2 curves with respect to different number of hash bits.}
	\label{fig:mnist-results}
\end{figure*}
\begin{figure*}[!htb]
	\centering
	\subfloat{\label{fig:nus-wide-results-map}\small{(a)}\raisebox{-.95\height}{\includegraphics[height=.205\textwidth]{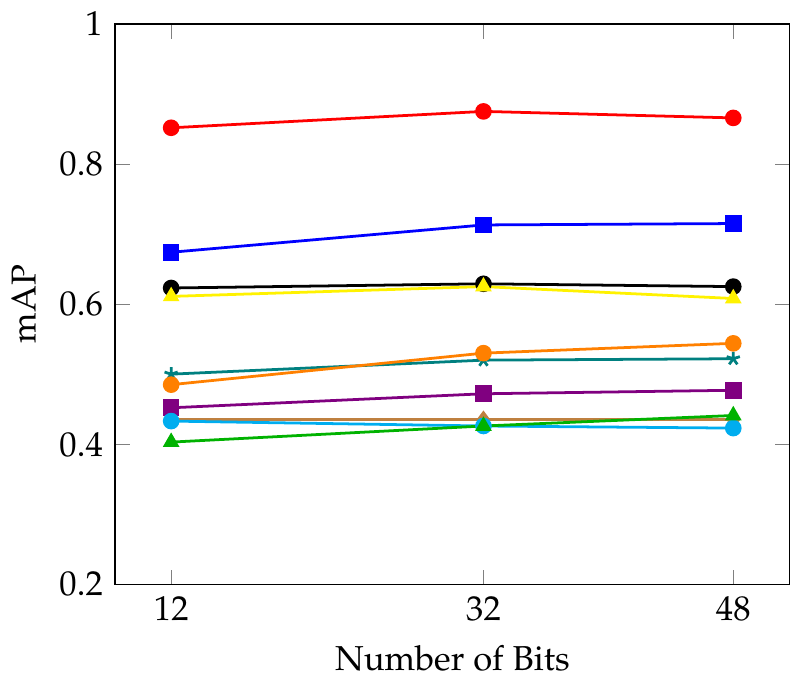}}}\qquad
	\subfloat{\label{fig:nus-wide-results-precision-k}\small{(b)}\raisebox{-.95\height}{\includegraphics[height=.205\textwidth]{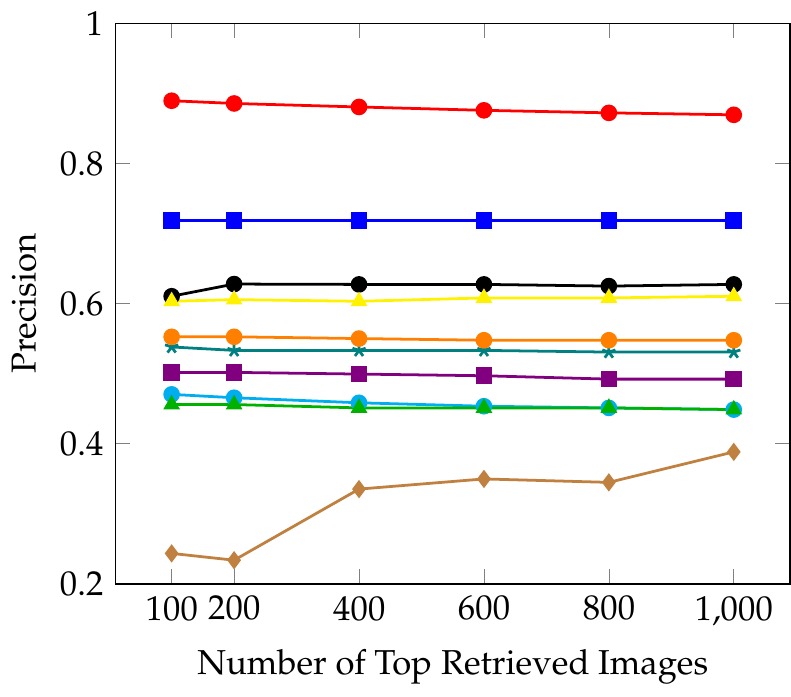}}}\qquad
	\subfloat{\label{fig:nus-wide-results-hammingradius}\small{(c)}\raisebox{-.95\height}{\includegraphics[height=.205\textwidth]{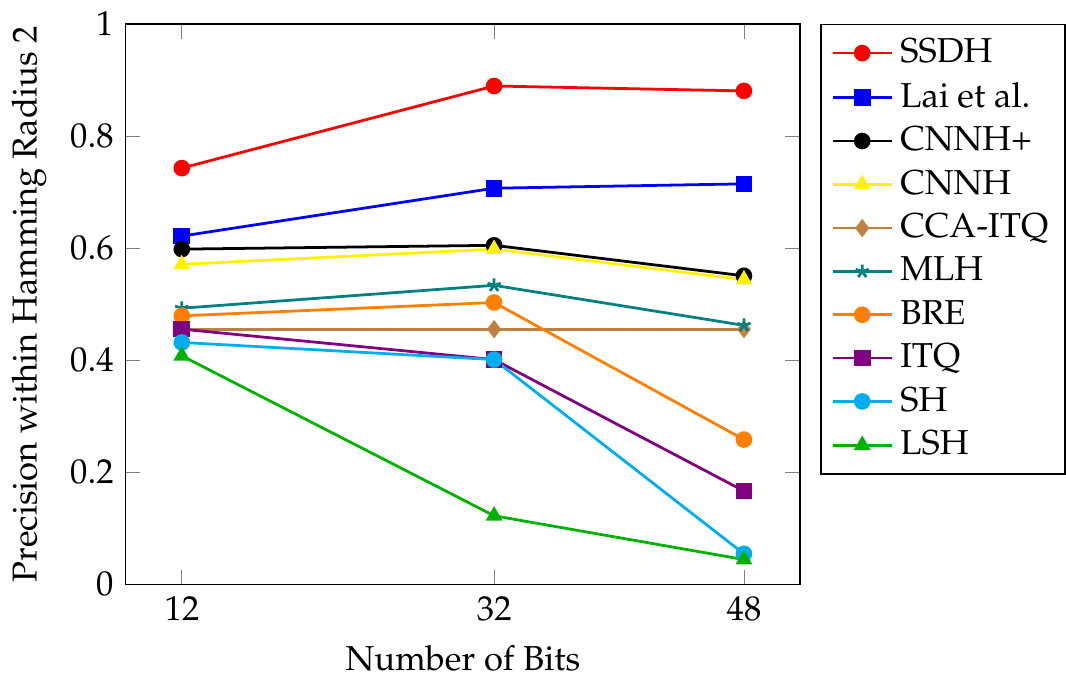}}}
	\caption{Comparative evaluation of different hashing algorithms on the NUS-WIDE dataset. (a) mAP curves of top 5,000 returned images with respect to different number of hash bits. (b) Precision curves with respect to different number of top retrieved samples when the 48-bit hash codes are used in the evaluation. (c) Precision within Hamming radius 2 curves with respect to different number of hash bits.}
	\label{fig:nus-wide-results}
\end{figure*}
\subsection{Retrieval Results on CIFAR-10}\label{subsec:cifar10-retrieval}

We compare SSDH with several hashing methods, including unsupervised methods (LSH~\cite{andoni:focs06}, ITQ~\cite{gong:pami13}, and SH~\cite{weiss:nips08}) and supervised approaches (BRE~\cite{kulis:nips09}, MLH~\cite{norouzi:icml11}, CCA-ITQ~\cite{gong:pami13}, CNNH+~\cite{xia:aaai14}, CNNH~\cite{xia:aaai14}, and Lai \etal~\cite{lai:cvpr15}).
\textcolor{black}{In the experiments, we use SSDH of the squared losses (i.e. $p=2$) in Eq.~\eqref{eqn:efficient-codes}, and the parameters $\alpha, \beta, \gamma$ in Eq.~\eqref{eqn:overall-objective} are all set as 1.}
Among the six supervised approaches, CNNH+, CNNH, and Lai \etal, like our approach, take advantage of deep learning techniques and supervised label information. 

Following the settings in~\cite{lai:cvpr15}, Figure~\ref{fig:cifar-results-map} shows the results based on the mAP as a function of code length.
Among various methods compared, it can be observed that the supervised approaches constantly outperform the unsupervised ones, LSH~\cite{andoni:focs06}, ITQ~\cite{gong:pami13} and SH~\cite{weiss:nips08}.
Besides, the deep learning-based approaches in~\cite{xia:aaai14, lai:cvpr15} and ours achieve relatively better performance, and this could be attributed to the fact that deep networks enable joint learning of feature representations and binary functions directly from images, and the learned feature representations are more effective than the hand-engineered ones such as 512-dimensional GIST features used in BRE~\cite{kulis:nips09}, MLH~\cite{norouzi:icml11}, and CCA-ITQ~\cite{gong:pami13}.

Referring to the results, SSDH provides stable and the most favorable performance for different code lengths, and improves the mAP by a margin of around 34\% compared with the competitive methods.
The results suggest that unifying retrieval and classification in a single learning model where the hash code learning is governed by the semantic labels can better capture the semantic information in images and hence yields more favorable performance.
Besides, compared to SDH~\cite{liong:cvpr15} that uses a different setting of 12-, 32-, and 64-bit codes that cannot be shown in the figure, the mAP obtained by our 12-bit SSDH is still much higher than 46.75\%, 51.01\%, and 52.50\%, respectively obtained in~\cite{liong:cvpr15}.

Figure~\ref{fig:cifar-results-precision-k} shows the precision at $k$ samples, where $k$ ranges from 100 to 1,000, when the 48-bit hash codes are used in the evaluation.
These curves convey similar messages as observed in the mAP measure. 
SSDH has a consistent advantage over other hashing methods, and the approaches (ours, Lai \etal, CNNH+, CNNH, and CCA-ITQ) that exploit the label information in learning hash functions perform better than those that do not.

The evaluation of the precision within Hamming radius 2 is shown in Figure~\ref{fig:cifar-results-hammingradius}.
Our approach performs more favorably against the others on this metric too.
As it is unclear what is the suitable value of $r$ for different tasks and code lengths, we consider the previous two evaluation metrics, mAP and precision at $k$ samples, would reflect the retrieval performance better than this metric in general.
Here, we use $r=2$ simply for following the conventions of performance comparison.

As our network is enhanced from a classification network, it is worth noting whether the classification performance is still maintained.
To verify this and for a fair comparison, we fine-tune the original AlexNet (\ie, the model without a latent layer added), initialized with the features trained on ImageNet, on the CIFAR-10 dataset.
The AlexNet+fine-tune achieves the classification accuracy of 89.28\% and our SSDH architecture (with a latent layer) attains the accuracies of 89.74\%, 89.87\% and 89.89\% for the code lengths 12, 32 and 48, respectively.
It reveals that stable classification performance is still accomplished by using our architecture.
More classification results for all of the single-labeled datasets can be found in Section~\ref{sec:classification}.

\begin{table*}[tb]
	\centering
	\setlength{\tabcolsep}{3pt}
	\caption{\textcolor{black}{The mAPs (\%) of SSDH with 48 bits versus $\beta$ and $\gamma$ while $\alpha$ is set to 1 on the CIFARI-10 dataset.}}
	\vspace*{-10pt}
	\label{tbl:term-combinations}
	\subfloat[Only $E_1$ and $E_2$ are applied]{%
	  \begin{tabular*}{0.23\textwidth}{@{}@{\extracolsep{\fill}}*{5}{r}@{}}
	  \toprule
	  \multicolumn{5}{c}{$\alpha = 1$ and $\gamma = 0$}\\
	  \midrule
	  $\beta = 0$ & 1 & 2 & 4 & 8\\
	  \midrule
	  90.70 & \textbf{91.19} & 91.14 & 90.50 & 90.24\\
	  \bottomrule
	  \end{tabular*}
	}\quad
	\subfloat[All three terms $E_1$, $E_2$, $E_3$ are applied and $\alpha$ is fixed to 1]{%
	  \begin{tabular*}{0.73\textwidth}{@{}@{\extracolsep{\fill}}*{15}{r}@{}}
	  \toprule
	  \multicolumn{5}{c}{$\beta = 0$} & \multicolumn{5}{c}{$\beta = 1$} & \multicolumn{5}{c}{$\beta = 2$}\\
	  \cmidrule{1-5}\cmidrule{6-10}\cmidrule{11-15}
	  $\gamma = 0$ & 1 & 2 & 4 & 8 & 0 & 1 & 2 & 4 & 8 & 0 & 1 & 2 & 4 & 8\\
	  \midrule
	  90.70 & 90.61 & 91.33 & 91.16 & 90.72 & 91.19 & \textbf{91.45} & 91.28 & 91.08 & 90.61 & 91.14 & 90.61 & 90.86 & 91.18 & 91.19\\
	  \bottomrule
	  \end{tabular*}
	}
\end{table*}

{\color{black}
We also study the influence of individual terms in the learning objective (with $p=2$ in Eq.~\eqref{eqn:efficient-codes}).
The loss of SSDH in Eq.~\eqref{eqn:overall-objective} consists of three terms encouraging label consistency, binarization, and equal sparsity of the codes.
First, we 
use only the two terms $E_1$ and $E_2$ by fixing the first weight $\alpha$ as 1, varying the second weight $\beta$ in $\{0, 2^0, 2^1, 2^2, 2^3$\}, and setting the third weight $\gamma$ as 0.
Table~\ref{tbl:term-combinations}a shows the mAPs of SSDH with 48-bit codes on the CIFAR-10 dataset.
It can be seen that the mAPs obtained are roughly around 90\%.
Among them, $\beta \in \{0, 2^0, 2^1\}$ get higher mAPs.
It reflects that a moderate level of binarization is helpful to binary codes learning.
We further study the case of adding the third term $E_3$ with $\alpha=1$, $\beta \in \{0, 2^0, 2^1\}$, and $\gamma \in \{0, 2^0, 2^1, 2^2, 2^3$\}, as shown in Table~\ref{tbl:term-combinations}b.
As can be seen, adding the equal-sparsity term ($E_3$) can possibly increase the performance too, and the equal weights $\alpha=\beta=\gamma=1$ get the highest mAP among all the situations studied.
Compare the cases where each term is getting added, $\{\alpha,\beta,\gamma\}=\{1,0,0\}$, $\{1,1,0\}$, and $\{1,1,1\}$.
The mAPs respectively obtained, 90.70\%, 91.19\%, and 91.45\%, are getting increased.
Hence, using all the terms is beneficial to achieving more favorable retrieval performance.
In the following, we simply choose the naive combination $\{\alpha,\beta,\gamma\}=\{1,1,1\}$ in Eq.~\eqref{eqn:overall-objective} for all of the other experiments and comparisons.
}
%
{\color{black}

Besides, we study the impacts of different functions on the performance by further using the L1-norm 
loss ($p=1$) in Eq.~\eqref{eqn:efficient-codes} and present empirical results in Table~\ref{tbl:comp-l1-l2-cifar10-mnist}.
We see that L1- and L2-norm losses attain comparable retrieval performance, indicating that our learning objective can provide stable results with different losses employed for learning binary codes.
Unless otherwise mentioned, we use $p=2$ in Eq.~\eqref{eqn:efficient-codes} in the following experiments.
}

\begin{table}[tb]
\setlength{\tabcolsep}{2pt}
	\centering
	\caption{\textcolor{black}{Performance comparison of using L1- and L2-losses in Eq.~\eqref{eqn:efficient-codes} on CIFAR-10 and MNIST based on mAP (\%).}}
	\label{tbl:comp-l1-l2-cifar10-mnist}
	\begin{tabular*}{.95\columnwidth}{@{}@{\extracolsep{\fill}}l*{6}{r}@{}}
	\toprule
	Loss     & \multicolumn{3}{c}{CIFAR-10} & \multicolumn{3}{c}{MNIST}\\
	         \cmidrule{2-4}\cmidrule{5-7}
\cmidrule{2-4}\cmidrule{5-7}
   & 12 & 32 & 48 & 12 & 32 & 48\\
	\midrule
	\textcolor{black}{$p=1$}       & 87.25   & 91.15   & 90.83  & 98.90 & 99.30 & 99.30\\
	\textcolor{black}{$p=2$}       & 90.59   & 90.63   & 91.45  & 99.31 & 99.37 & 99.39\\
	\bottomrule	
	\end{tabular*}
\end{table}
\subsection{Retrieval Results on MNIST}
MNIST is a relatively simpler dataset than CIFAR10.
Though many methods can get fairly good performance on the dataset, we show that the performance can still be improved by SSDH.
Figure~\ref{fig:mnist-results} shows the comparison of different hashing methods on MNIST.
We see that these results accord with our observations in CIFAR-10 that SSDH performs more favorably against other supervised and unsupervised approaches. 

We also report the classification performance for this single-labeled dataset.
The AlexNet+fine-tune achieves the classification accuracy of 99.39\% and our SSDH achieves 99.40\%, 99.34\% and 99.33\% for the code lengths 12, 32 and 48, respectively.
This shows again that our architecture can retain similar performance for the classification task under the situation that lower dimensional features (from 4096-d to 12/32/48-d) are extracted.

{\color{black}
Besides, following CIFAR-10, we also study the effects of different loss functions in Eq.~\eqref{eqn:efficient-codes}.
The results reported in Table~\ref{tbl:comp-l1-l2-cifar10-mnist} show that the performance of $p=1$ is on a par with that of $p=2$, confirming again that both L1- and L2-norms in Eq.~\eqref{eqn:efficient-codes} are capable of learning good codes.}
\subsection{Retrieval Results on NUS-WIDE}
SSDH is also compared with several unsupervised and supervised approaches on NUS-WIDE, similar to the evaluation done on CIFAR-10.
As the web images in NUS-WIDE are associated with more than one label, SSDH is trained to optimize the proposed maximum-margin loss \textcolor{black}{in Eq.~\eqref{eqn:multi-label_loss}} for classification along with the two other terms for efficient binary codes construction.

Following also the settings of \cite{lai:cvpr15}, the comparisons of various approaches are shown in Figure~\ref{fig:nus-wide-results}, where the relevance of the retrieved image and the query image is verified by whether they share at least one common label.
Like the results in CIFAR-10 and MNIST, the performance of supervised and deep approaches are better than non-supervised and non-deep approaches.
Our SSDH produces constantly better results than the other approaches when the performance is evaluated according to the mAP of top 5,000 returned images and the precision at $k$ samples for $k=100$ to 1,000.
The improvement SSDH obtains over the previous state-of-the-art results in mAPs is about 16\% (Figure~\ref{fig:nus-wide-results}a) and in precision at $k$ samples (Figure~\ref{fig:nus-wide-results}b) is about 16\%.

When evaluated by the precision within Hamming radius 2, SSDH also provides better results.
As discussed in the results of CIFAR-10, this 
metric would not reflect the performance properly when the code length is long.
As can be seen, the performance on this metric drops for longer codes in our method, which could reflect that our method can balance the semantic information captured by the bits.

In sum, the results are consistent with those of CIFAR-10 and MNIST, suggesting that SSDH is a general network that can deal with images associated with multiple labels or with a single label.
{\color{black}
We also study the impact of using L1 margin ($p=1$) in implementing the maximum-margin loss of Eq.~\eqref{eqn:multi-label_loss}.
The comparison 
in Table~\ref{tbl:comp-l1-l2-nuswide} indicates that the retrieval performance of L2 margin is greatly better than that of L1 margin.
This would be because the gradients in L2 margin depend on the distances between misclassified samples to the true labels, allowing a network to easily correct misclassified samples, but the gradients of L1 margin (either 1 or $-1$) are irrespective of the distances between them, perhaps leading to inferior performance.
Note that though using L1 margin degrades the performance, our approach still obtains better results than the previously competitive 
method~\cite{lai:cvpr15} that achieves mAPs of 67.4\%, 71.3\%, and 71.5\% for 12, 32, and 48 bits, respectively.
}
\begin{table}[tb]
	\centering
	\caption{\textcolor{black}{Performance comparison of using L1- and L2-margin losses in Eq.~\eqref{eqn:multi-label_loss} on NUS-WIDE based on mAP (\%) and precision (\%) at 500 samples.}}
	\label{tbl:comp-l1-l2-nuswide}
	\begin{tabular*}{.95\columnwidth}{@{}@{\extracolsep{\fill}}lrrrrrr@{}}
	\toprule
	Loss 	& \multicolumn{3}{c}{mAP (\%)} & \multicolumn{3}{c}{prec. (\%) @ 500}\\
		\cmidrule{2-4}\cmidrule{5-7}
	         & 12 & 32 & 48  & 12 & 32 & 48\\
	\midrule
	\textcolor{black}{$p=1$}       & 71.73   & 82.85   & 83.97 & 71.70 & 84.37 & 85.50\\
	\textcolor{black}{$p=2$}       & 85.17   & 87.51   & 86.58 & 87.64 & 89.05 & 87.83\\
	\bottomrule	
	\end{tabular*}
\end{table}

\begin{figure}[!t]
	\centering
	\includegraphics[width=.78\columnwidth]{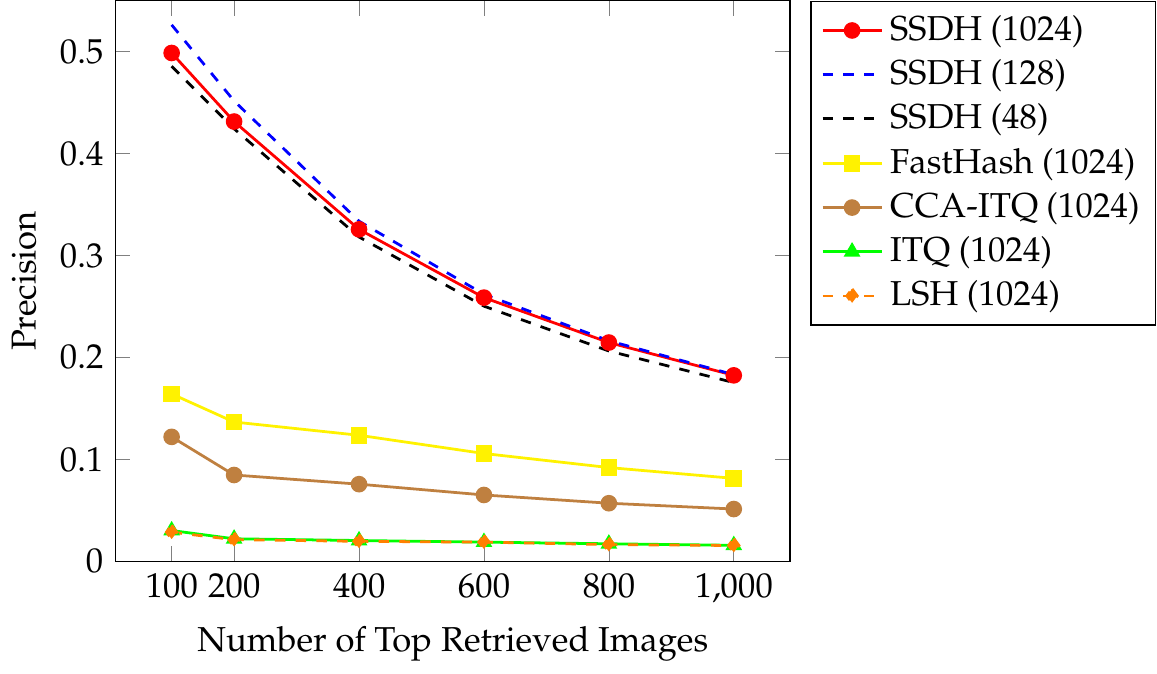}
	\caption{Precision curves with respect to different number of top retrieved samples on the SUN397 dataset. The number inside parentheses indicates the code length.}
	\label{fig:sun-results}
\end{figure}

\subsection{Retrieval Results on SUN397}
SUN397 comprises more than 100,000 images in 397 scene categories.
It is more challenging than CIFAR-10 and MNIST. 
Following the settings in~\cite{lin:cvpr14}, we choose the code length as 1024 bits for comparison.
Figure~\ref{fig:sun-results} compares SSDH, FastHash~\cite{lin:cvpr14}, CCA-ITQ, ITQ, and LSH based on the precision at different number of top returned images.
SSDH performs better than the other approaches regardless of the number of top returned images.
In addition, the advantage is more remarkable when a small number of top returned images are needed.
When only the top 200 returned images are considered, SSDH outperforms FastHash by a margin of 30\% precision.
Thus, even for the case when code sizes are large, SSDH achieves state-of-the-art hash-based retrieval performance.
We also apply SSDH to the dataset when the code lengths are 128 and 48 bits and obtain precision curves close to that of SSDH with 1024 bits.
The result shows that the performance of our approach still keeps good even when the codes are far 
\textcolor{black}{shorter than the number of classes, 397}.

The results are obtained using the pre-trained weights on ImageNet that contains object-based images.
Because SUN397 contains mainly scene-based images, the performance is likely to be boosted by using the initial weights pre-trained on another big dataset, Places dataset~\cite{zhou:nips14}.
However, to coincide with the other experiments, we report the results initialized by the ImageNet pre-trained weights here.
We also implement the fine-tuned AlexNet for the comparison of the classification performance.
The fine-tuned AlexNet achieves a classification accuracy of 52.53\% that is moderately better than the result (42.61\%) reported in~\cite{zhou:nips14} which uses AlexNet features without fine-tuning.
Our SSDH achieves classification accuracies of 53.86\%, 53.24\% and 49.55\% when code lengths are 1024, 128, and 48, respectively, revealing again that the classification performance is 
maintained in our architectural enhancement.
\begin{figure}[!t]
	\centering
	\includegraphics[width=.78\columnwidth]{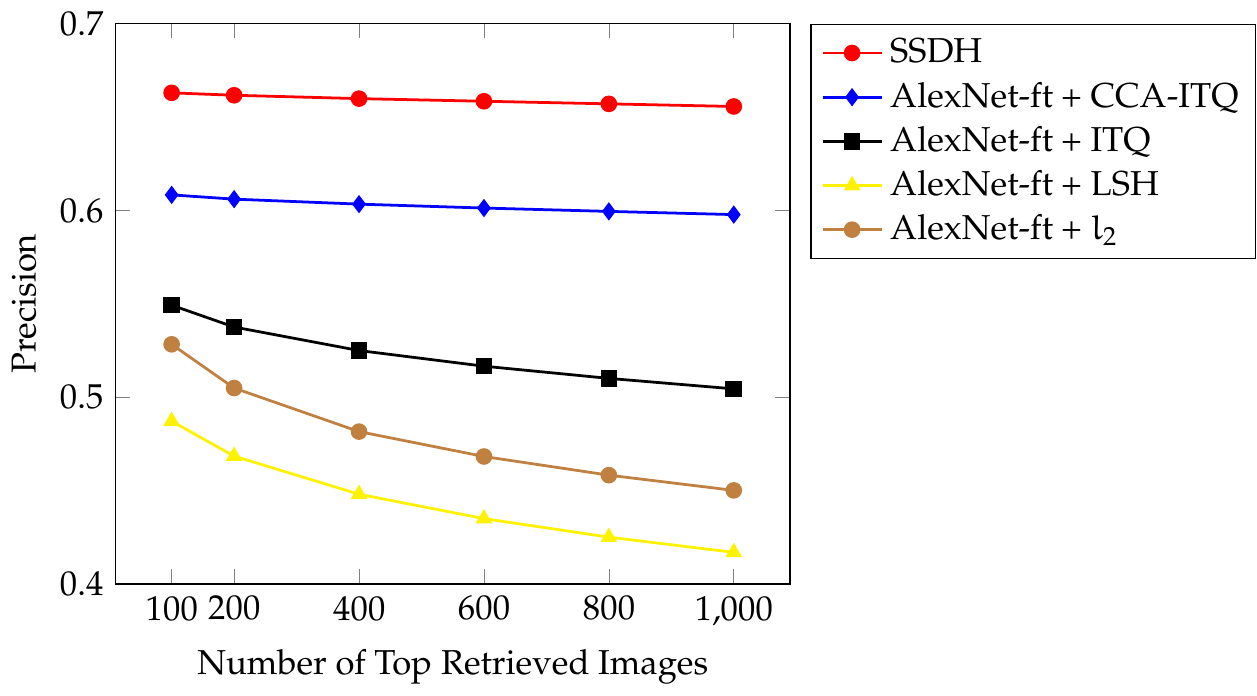}
	\caption{Precision curves with respect to different number of top retrieved samples on the Yahoo-1M dataset when the 128-bit hash codes are used in the evaluation. AlexNet-ft denotes that the features from layer $F_7$ of AlexNet fine-tuned on Yahoo-1M are used in learning hash codes.}
	\label{fig:yahoo-results}
\end{figure}
\begin{table}[!t]
	\centering
	\caption{mAP (\%) of various methods at 128 bits on the Yahoo-1M dataset. AlexNet-ft denotes that the features from layer $F_7$ of AlexNet fine-tuned on Yahoo-1M are used in learning hash codes.}
	\label{tbl:yahoo-map}
	\begin{tabular*}{.85\columnwidth}{@{}@{\extracolsep{\fill}}lr@{}}
	\toprule
	Method     & mAP \\
	\midrule
	AlexNet-ft + $l_2$   & 48.95\\
	AlexNet-ft + LSH     & 46.39\\
	AlexNet-ft + ITQ     & 53.86\\
	AlexNet-ft + CCA-ITQ & 61.69\\
	\midrule
	SSDH                 & 66.63\\
	\bottomrule
	\end{tabular*}
\end{table}
\subsection{Retrieval Results on Yahoo-1M Dataset}\label{subsec:yahoo-retrieval}
Yahoo-1M is a single-labeled large-scale dataset.
Hashing approaches that require pair- or triple-wised inputs for learning binary codes are unsuitable for end-to-end learning on Yahoo-1M due to the large time and storage complexities.
We hence compare SSDH with point-wised methods that are applicable to such a large dataset.
We fine-tune AlexNet on Yahoo-1M and then apply LSH, ITQ, and CCA-ITQ to learn the hash codes from the layer $F_7$ features.
These two-stage (AlexNet fine-tune+X) approaches serve as the baselines compared in this experiment.
To provide more insight into the performance of the hash approaches, we also include the results obtained by the Euclidean ($l_2$) distance of the $F_7$ features from the fine-tuned AlexNet in the comparison.
The hash approaches are evaluated when the code length is 128.

Figure~\ref{fig:yahoo-results} shows the precision curves with respect to a different number of top retrieved images and Table~\ref{tbl:yahoo-map} shows the mAP of the top 1,000 returned images.
We compute the mAP based on the top 1,000 images of a returned list rather than the entire list due to the high computational cost in mAP evaluation.
It is interesting that the hash approaches, except LSH, give better retrieval performance than a direct match based on the Euclidean distance of the fine-tuned deep features.
This shows that learning hash codes on top of the deep features can improve the quantization in the feature space and increase the retrieval performance.
The results also show that supervised hashing approaches can better capture the semantic structure of the data than unsupervised ones.
Furthermore, SSDH gets more favorable results than the two-stage approaches combining fine-tuned AlexNet features and conventional hash methods.
We owe this to an advantage of our approach that simultaneous learning of the deep features and hash functions can achieve better performance.
About the classification performance, SSDH and fine-tuned AlexNet get 73.27\% and 71.86\% accuracies, respectively.
\begin{figure}[tb]
	\centering
	\includegraphics[width=.78\columnwidth]{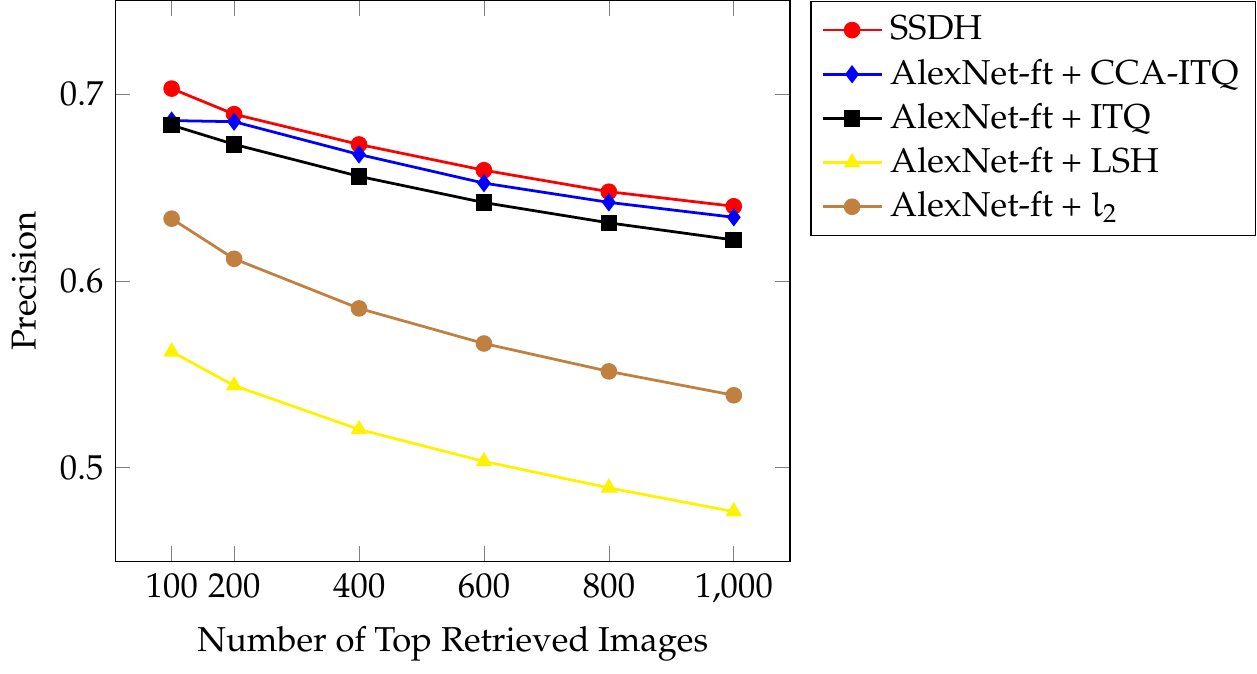}
	\caption{Precision curves with respect to different number of top retrieved samples on the UT-ZAP50K dataset when the 48-bit hash codes are used in the evaluation. AlexNet-ft denotes that the features from layer $F_7$ of AlexNet fine-tuned on UT-ZAP50K are used in learning hash codes.}
	\label{fig:utzap50k-results}
\end{figure}
\subsection{Retrieval Results on UT-ZAP50K}\label{subsec:utzap50k-retrieval}
UT-ZAP50K is a multi-label dataset consisting of shopping images, which has not been used for retrieval performance comparison yet.
Similar to the experiments on Yahoo-1M, we use deep features from fine-tuned AlexNet for LSH, ITQ, and CCA-ITQ to learn binary codes and also include the performance of an exhaustive search based on the Euclidean ($l_2$) distance of the deep AlexNet features.
The performance is evaluated when the code length is 48.

In this experiment, we verify the relevance of the query and returned images by examining whether they have exactly the same labels.
This is because when searching shopping items, one may want the retrieved images not only in the same category but also for the same gender to the query.
This criterion requires all relevant labels to be retrieved for a query, which is stricter than that for the NUS-WIDE dataset where the retrieval is considered correct if it exhibits at least one common labels with the query.

Figure~\ref{fig:utzap50k-results} shows the precision of various methods at top $k$ returned images.
Under such a demanding evaluation criterion, SSDH still produces better results than the compared approaches for all $k$.
Similar to the results of Yahoo-1M, the hash-based approaches (AlexNet-FineTune+ITQ, AlexNet-FineTune+CCA-ITQ, and ours) can yield effective quantization spaces and get more favorable results than searching with fine-tuned AlexNet features in Euclidean space.

{\color{black}
Like NUS-WIDE, we investigate the use of L1 margin ($p=1$) in the maximum-margin loss of Eq.~\eqref{eqn:multi-label_loss} for this multi-label dataset.
When implemented with 48-bit codes, SSDH produces a 65.94\% mAP and a 62.08\% precision@500 samples.
These results are worse than the 71.91\% mAP and the 66.59\% precision@500 samples of SSDH with L2 margin, in accordance with the observations made on NUS-WIDE.
Hence, from these results, we suggest to use $p=2$ in the maximum-margin loss in Eq.~\eqref{eqn:multi-label_loss} for multi-label learning.
}

{\color{black}
\subsection{Retrieval Results on ILSVRC2012}\label{subsec:ilsvrc12-retrieval}

Thus far, the number of dataset labels having been handled is around 10 to 100, except that SUN397 has approximately 400 labels.
In this experiment, we apply SSDH to the ILSVRC2012 dataset that is large in both data amount and number of labels to further demonstrate the scalability of SSDH.
We compare SSDH with the combinations of AlexNet features and ITQ/CCA-ITQ because they perform considerably better than AlexNet-FineTune+$l_2$ and AlexNet-FineTune+LSH on the Yahoo-1M and UT-ZAP50K datasets.
Since the AlexNet model (from CAFFE) has been pre-trained on this dataset, we directly use the AlexNet features extracted as the input for ITQ and CCA-ITQ.
Besides, as ITQ and CCA-ITQ require high memory usage for matrix computation, only 100,000 samples 
are deployed for the subspace learning of them.
For our SSDH, a 512-bit latent layer is used and our SSDH is then fine-tuned on ILSVRC2012.

The upper half of Table~\ref{tbl:ilsvrc2012} shows the results. 
It appears that SSDH constantly yields better performance, 
which 
confirms that SSDH 
is applicable to not only large datasets but also the data of numerous and diverse labels.

\begin{table}[tb]
	\centering
	\setlength{\tabcolsep}{0pt}
	\caption{\textcolor{black}{The mAP at top 1,000 returned images and precision at $k$ samples of methods on the ILSVRC2012 validation set. The code size is 512.}}
	\label{tbl:ilsvrc2012}
	\setlength{\tabcolsep}{0pt}
	\begin{tabular*}{\columnwidth}{@{}@{\extracolsep{\fill}}lrrrrrr@{}}
	\toprule
	Method  & mAP (\%)  & \multicolumn{5}{c}{prec. (\%) at $k$ samples}  \\
	\cmidrule{3-7}
	        &       & 200   & 400   & 600   & 800   & 1,000\\
	\midrule
	\textcolor{black}{AlexNet + ITQ}     & 31.21 & 32.23 & 28.54 & 25.82 & 23.59 & 21.69\\
	\textcolor{black}{AlexNet + CCA-ITQ} & 38.03 & 39.10 & 36.64 & 34.48 & 32.37 & 30.25\\	
	\textcolor{black}{SSDH, AlexNet}    & 46.07 & 47.27 & 45.59 & 43.76 & 41.65 & 39.23\\
	\midrule
	\textcolor{black}{VGG16 + ITQ}     & 47.07 & 49.00 & 45.30 & 42.10 & 39.09 & 36.17\\
	\textcolor{black}{VGG16 + CCA-ITQ} & 52.74 & 53.91 & 51.68 & 49.56 & 47.28 & 44.68\\
	\textcolor{black}{SSDH, VGG16}     & 61.47 & 62.88 & 61.22 & 59.40 & 57.19 & 54.41\\
	\bottomrule
	\end{tabular*}
\end{table}
\subsection{Retrieval Using Different Networks}\label{sec:different-nets}


Our SSDH can be generally integrated with other networks.
In this section, we provide the retrieval results of SSDH with VGG16 (configuration D in \cite{simonyan:iclr15}), aside from AlexNet.
VGG16 is much 
deeper than AlexNet.
It comprises 13 convolutional layers followed by 2 fully connected and one output layers, and small (\eg, $3 \times 3$) convolution filters are exploited. 
Like the way of applying our SSDH to AlexNet, a latent layer is added between the output layer and its previous layer in VGG16 to realize our approach. 

Table~\ref{tbl:comp-vgg-alexnet} shows the results on CIFAR-10, NUS-WIDE, Yahoo-1M, and ILSVRC2012. 
For the large-scale datasets, Yahoo-1M and ILSVRC2012, we observe that VGG16 can boost SSDH's performance by an at least 8.8\% higher mAP 
Therefore, deeper networks can learn more effective binary representations from complex and large-scale data.
For small- (CIFAR-10) and medium-sized (NUS-WIDE) datasets, SSDH with both networks attain similar performance, reflecting that a less complex network should suffice for handling small-sized data.
These results reveal that SSDH can be established on different architectures for the applications of different data sizes.
In addition, the characteristic of its capability of leveraging on existing networks also makes it easily implementable and flexible for practical use.



\noindent \textbf{Network simplification.} To benefit large-scale image search, 
fast hash code computation is required.
Thus, an interesting question arises.
Can other network configurations allow for fast code computation and also provide comparable results?
\begin{table}[tb]
	\centering
	\caption{\textcolor{black}{The mAPs of SSDH with different deep models on CIFAR-10, NUS-WIDE, Yahoo-1M, and ILSVRC2012.
}}
	\label{tbl:comp-vgg-alexnet}
	\setlength{\tabcolsep}{0pt}
	\begin{tabular*}{.95\columnwidth}{@{}@{\extracolsep{\fill}}lrrrr@{}}
	\toprule
	Method   & CIFAR-10           & NUS-WIDE & Yahoo-1M & ILSVRC2012\\
	         \cmidrule{2-2}\cmidrule{3-3}\cmidrule{4-4}\cmidrule{5-5}
	         & 48            & 48      & 128    & 512\\
	\midrule
	SSDH, AlexNet  & 91.45   & 86.58   & 66.63  & 46.07\\
	SSDH, VGG16    & 92.69   & 88.97   & 75.45  & 61.47\\
	\bottomrule	
	\end{tabular*}
\end{table}
To address this issue, we conduct experiments with two more networks, VGG11 (configuration A in \cite{simonyan:iclr15}) and VGG-Avg (of our own design),
on the CIFAR-10 dataset.
\begin{itemize}[leftmargin=*]
\item VGG11~\cite{simonyan:iclr15} is similar to 
VGG16.
They differ only in depth: VGG11 has 11 layers (8 convolutional, 2 fully connected, and one output layers), whereas VGG16 has 16 layers.
\item VGG-Avg is modified from VGG16 by ourselves. It comprises the same 13 convolutional layers as VGG16, but the fully-connected layers (with the output classification layer excluded) in VGG16 are replaced by an average pooling layer. 
Because the last convolutional layer of VGG16 has 512 channels, the average pooling 
produces a 512-dimensional feature vector.
This vector is then connected to a 48-bit latent layer followed by a final classification layer in our SSDH.
The design is inspired by the counterpart of NIN~\cite{lin:iclr14} and the very new and successful ResNet~\cite{he:cvpr16}.
It decreases the number of network parameters drastically ---
89\% out of the VGG16's 134 M parameters are taken up by the fully connected layers, while no parameters need to be learned for average pooling.
The model size of VGG-Avg (15~M) is even smaller than that of VGG11 (129~M) and AlexNet (57~M) as shown in Table~\ref{tbl:model-size}, making it a cheaper network consuming less resources.
Because the average pooling preserves the shift invariance of the convolutional layers, the extracted features are still effective for classifying an entire image.

\end{itemize}
The mAPs of SSDH with VGG11, VGG16, VGG-Avg and AlexNet are 88.40\%, 92.69\%, 90.75\% and 91.45\% on the standard benchmark CIFAR-10, respectively, where VGG11 performs less favorably.
We conjecture that fewer layers combined with small-sized filters limits its ability to learn representative codes.
VGG-Avg performs better than VGG11 (though slightly worse than VGG16), revealing that replacing the fully connected layers by average pooling highly reduces the network complexity with only a little drop on the retrieval performance.

\begin{table}[tb]
	\centering
	\caption{\textcolor{black}{Number of parameters and amount of storage of different network models with a 48-bit latent layer (in CAFFE).}}
\label{tbl:model-size}
	\begin{tabular*}{.95\columnwidth}{@{}@{\extracolsep{\fill}}lrrrr@{}}
	\toprule
				  & \multicolumn{4}{c}{SSDH-48}\\
	\cmidrule{2-5}
	              & AlexNet & VGG16 & VGG11 & VGG-Avg\\
	\midrule	
	\# parameters & 57 M    & 134 M & 129 M & 15 M\\
	required storage & 228 MB & 537 MB & 516 MB & 59 MB\\
	\bottomrule
	\end{tabular*}
\end{table}

\subsection{Cross Domain and Label Learning}\label{sec:transfer}
We now study the 
usage of SSDH in two aspects: (1) cross-domain retrieval, \ie, trained on
one dataset and applied to another and (2) retrieval on datasets with missing labels.


\noindent \textbf{Cross-domain instance-level retrieval.}
SSDH is a supervised hash method.
It uses the image labels in the training dataset (i.e., gallery)  to learn compact binary codes.
Each image in the gallery is then given with a binary code that can be pre-stored for fast retrieval.
However, typical instance-level datasets such as Paris and Oxford lack such semantic-label annotations.
Their image relevancies are mainly established by near-duplicates.

We use our SSDH on these datasets to examine its capability in similarity-based image retrieval.
The centerpiece of SSDH is established on the idea that semantic label classification is driven by several latent binary attributes; semantic labels are thus needed in SSDH training.
To apply SSDH to both datasets without labels, we follow the idea of neural codes for image retrieval~\cite{babenko:eccv14} that
the SSDH network is pre-trained on a related dataset with label supervision.
This pre-trained dataset, Landmarks~\cite{babenko:eccv14}, contains URLs of 270,000+ images.
Following the given URLs, we were able to download 214,141 images of 721 labels.
The SSDH of VGG16 is used to learn a network model from the downloaded dataset, where a 512-bit latent layer is used because of its better performance on large-scale datasets.
We then use the network model to extract binary codes for Paris and Oxford datasets without any further fine-tuning.

The Paris and Oxford pose a challenge to instance-level retrieval as the same object in different images may appear
in distinct viewpoints and scales.
Similarities between images may thus be determined by some local patches.
To deal with this issue, we follow the spatial search~\cite{razavian:cvprws14} approach, where the image
relevance is established based on our binary codes of local patches at multiple scales.
The distance between a local query patch and a gallery image is defined as the minimum among the Hamming distances
of that query and gallery patch pairs.
Then, the average Hamming distance of all query patches to the gallery is used as the distance between the query and
the gallery.

Table~\ref{tbl:instance-level-results} compares our retrieval results with the others, where we are one of the few providing results based on binary hash codes for instance-level retrieval.
Among the other results, only the one in~\cite{morere:cbmm06} is based on binary codes of 512 bits; the rest rely on
real-valued features of 256~\cite{babenko:iccv15}, 512~\cite{babenko:eccv14}, or higher than
4,096~\cite{ng:cvprws15,razavian:cvprws14} dimensions, and all methods take advantage of deep
learning techniques.
For Paris that is a dataset with a moderate level of viewpoint and scale changes, SSDH 
performs more favorably against the other approaches.
For Oxford that is a dataset with stronger viewpoint and scale changes, SSDH performs not the best but is still
competitive.
Nevertheless, SSDH achieves the performance by using a more compact code (512-bit) than the others that use
real-valued codes.
Compared with the approach using binary codes of the same length \cite{morere:cbmm06}, SSDH still performs more favorably.
The results show that the models trained on a large dataset can be applied to the tasks in a relevant domain.
Besides, the outcomes also reveal that the codes learned 
are applicable to retrieval tasks in which visual similarity is the criterion to determine the relevance between
images.



\begin{table}[tb]
	\centering
	\caption{\textcolor{black}{Comparison of the instance-level retrieval performance (mAP (\%)) of SSDH with other
approaches on the Paris and Oxford datasets.}}
	\label{tbl:instance-level-results}
	\begin{tabular*}{.85\columnwidth}{@{}@{\extracolsep{\fill}}lrr@{}}
	\toprule
	Method   & Paris & Oxford          \\
	\midrule
	Neural codes~\cite{babenko:eccv14} & ---   & 55.70\\
	Ng \etal~\cite{ng:cvprws15}        & 69.40 & 64.90 \\
	CNN-aug-ss~\cite{razavian:cvprws14}& 79.50 & 68.00\\
	Sum pooling~\cite{babenko:iccv15}  & ---   & 58.90\\
	Mor{\`e}re \etal~\cite{morere:cbmm06}, 512 bits              & ---   & 52.30\\
	\midrule
	SSDH w/ 512-bit codes, spatial search    & 83.87 & 63.79   \\
	\bottomrule	
	\end{tabular*}
\end{table}

\noindent \textbf{Retrieval on datasets with missing labels.} In this experiment, we consider the setting that learning is performed on a dataset with missing labels.
We choose the multi-label dataset, NUS-WIDE, for the evaluation.
For each training image with more than one label in NUS-WIDE, half of its labels are randomly removed.
In this way, about 55\% of the training images have 50\% missing labels, and the testing set remains the same with complete labels.
To handle the missing labels in the implementation, we treat them as ``don't care'' in CAFFE.
That is, the missing labels do not contribute to the error terms in the classification layer during training.
SSDH of the code length 48 with the VGG16 model is used in this experiment.

The results are reported as follows.
On the missing-labels setting, SSDH still gets an mAP of 88.02\%, only a slight drop from the 88.97\% of the complete-labels setting shown in Table~\ref{tbl:comp-vgg-alexnet}.
This indicates that SSDH can learn effective models from the cross-label information in a multi-label dataset, and performs robustly under label missing.
}
\subsection{Computational Time}
One advantage that binary codes offer is faster code comparison.
For instance, it takes about 51.83~$\micro$s to compute the Euclidean distance of two 4096-d floating-point features with a MATLAB implementation on a desktop with an Intel Xeon 3.70 GHz CPU of 4 cores, yet comparing two 512-bit (128-bit) binary codes takes only about 0.17~$\micro$s (0.04~$\micro$s).


\subsection{Classification Results on Various Datasets}\label{sec:classification}
In previous sections, we have depicted the classification performance 
of SSDH 
for the single-labeled datasets.
In this section, we present more classification results on the benchmark datasets in \textcolor{black}{Table~\ref{tbl:all-datasets-classification}}.
From the table, it is observed that our approach yields comparable performance to the state-of-the-art classification accuracies.
An interesting finding is that our approach achieves close 
classification accuracies compared to the fine-tuned AlexNet \textcolor{black}{or VGG}.
In particular, the performance is attained via a rather lower-dimensional feature space (eg. a 48-, 128-, or 512-dimensional binary feature space) that is more compact, while the AlexNet \textcolor{black}{or VGG} feature is of 4096 dimension of real values.
Because the classification task relies on the feature space learned, it thus shows that our architecture can cast the input image into a considerably lower-dimensional space with an approximate class separation capability for the same data.
The outcomes suggest that SSDH, a multi-purpose architecture for retrieval and classification, not only achieves promising classification performance when compared with the models that are optimized for a classification task, but also is beneficial to the retrieval task. 

Some further remarks and discussions of the experimental results are given in Appendix A.



\begin{table}[!t]
	\centering
	\caption{\textcolor{black}{Classification accuracy of various methods on CIFAR-10, 
SUN397, Yahoo-1M, and ILSVRC2012.} }
	\label{tbl:all-datasets-classification}
	\begin{tabular*}{.85\columnwidth}{@{}@{\extracolsep{\fill}}lrr@{}}
	\toprule
	Dataset, Method                      & \multicolumn{2}{c}{Accuracy (\%)}\\
	\midrule
	CIFAR-10\\
	\qquad Stochastic Pooling~\cite{zeiler:iclr13}      &  & 84.87\\
	\qquad CNN + Spearmint~\cite{snoek:nips12}          &  & 85.02\\
	\qquad NIN + Dropout~\cite{lin:iclr14}              &  & 89.59\\
	\qquad NIN + Dropout + Augmentation~\cite{lin:iclr14}& & 91.19\\
	\qquad AlexNet + Fine-tuning                         & & 89.28\\
	\qquad SSDH w/ 12-bit codes, AlexNet        & & 89.74\\ 
	\qquad SSDH w/ 32-bit codes, AlexNet        & & 89.87\\ 
	\qquad SSDH w/ 48-bit codes, AlexNet        & & 89.89\\
	\qquad \textcolor{black}{SSDH w/ 48-bit codes, VGG16}  & & 91.51\\
	\qquad \textcolor{black}{SSDH w/ 48-bit codes, VGG11}  & & 85.99\\
	\qquad \textcolor{black}{SSDH w/ 48-bit codes, VGG-Avg}  & & 90.54\\
	\midrule
	SUN397\\
    \qquad Cascade fine-tuned CNN~\cite{jie:accv14}       & &46.87\\
	\qquad MOP-CNN~\cite{gong:eccv14}                     & &51.98\\
	\qquad AlexNet + Fine-tuning                          & &52.53\\
	\qquad SSDH w/ 48-bit codes, AlexNet            & &49.55\\ 
	\qquad SSDH w/ 128-bit codes, AlexNet           & &53.24\\ 
	\qquad SSDH w/ 1024-bit codes, AlexNet          & &53.86\\ 
	\qquad \textcolor{black}{VGG16 + Fine-tuning}                     & & 64.68\\
	\qquad \textcolor{black}{SSDH w/ 128-bit codes, VGG16}             & & 61.54\\
	\midrule
	Yahoo-1M\\
	\qquad AlexNet + Fine-tuning                          &  &71.86\\
	\qquad SSDH w/ 128-bit codes, AlexNet         &  &73.27\\
    \qquad \textcolor{black}{SSDH w/ 128-bit codes, VGG16}        &  &78.86\\
    \midrule
    \textcolor{black}{ILSVRC2012}    & top-5  & top-1\\
    	\qquad \textcolor{black}{Overfeat~\cite{sermanet:iclr14}} & 85.82  & 64.26 \\
	\qquad \textcolor{black}{AlexNet}                  	 & 80.03 & 56.90  \\
	\qquad \textcolor{black}{SSDH w/ 512-bit codes, AlexNet}&78.69      & 55.16 \\
	\qquad \textcolor{black}{VGG16}                        & 88.37& 68.28  \\
	\qquad \textcolor{black}{SSDH w/ 512-bit codes, VGG16} & 89.76       &  70.51 \\
    \qquad \textcolor{black}{SSDH w/ 1024-bit codes, VGG16} &90.19  &  71.02 \\
	\bottomrule	
	\end{tabular*}
\end{table}
\vspace*{-5pt}

\section{Conclusions}\label{sec:conclusion}
%
%

We have presented a supervised deep hashing model, SSDH, that preserves the label semantics between images.
SSDH constructs hash functions as a latent layer between the feature layer and the classification layer in a network.
By optimizing an objective function defined over classification error and desired criterion for binary codes, SSDH jointly learn binary codes, features, and classification.
Such a network design comes with several merits: (1) SSDH unifies retrieval and classification in a single model; and (2) SSDH is simple and is easily realized by a slight modification of an existing deep network for classification; and (3) SSDH is naturally scalable to large scale search.
We have conducted extensive experiments and have provided comparative evaluation of SSDH with several state-of-the-arts on many benchmarks with a wide range of image types.
The results have shown that SSDH achieves superior retrieval performance and provides promising classification results. 
\vspace*{-5pt}
\ifCLASSOPTIONcaptionsoff
  \newpage
\fi

  \section*{Acknowledgment}
%
%
We thank the anonymous reviewers for their insightful comments. This work is supported in part by the Ministry of Science and Technology of Taiwan under contract MOST 104-2221-E-001-023-MY2 and MOST 105-2218-E-001-006.



\bibliographystyle{IEEEtran}
\bibliography{./bib/hashing,./bib/deep-learning,./bib/retrieval}

\begin{thebibliography}{10}
\providecommand{\url}[1]{#1}
\csname url@samestyle\endcsname
\providecommand{\newblock}{\relax}
\providecommand{\bibinfo}[2]{#2}
\providecommand{\BIBentrySTDinterwordspacing}{\spaceskip=0pt\relax}
\providecommand{\BIBentryALTinterwordstretchfactor}{4}
\providecommand{\BIBentryALTinterwordspacing}{\spaceskip=\fontdimen2\font plus
\BIBentryALTinterwordstretchfactor\fontdimen3\font minus
  \fontdimen4\font\relax}
\providecommand{\BIBforeignlanguage}[2]{{%
\expandafter\ifx\csname l@#1\endcsname\relax
\typeout{** WARNING: IEEEtran.bst: No hyphenation pattern has been}%
\typeout{** loaded for the language `#1'. Using the pattern for}%
\typeout{** the default language instead.}%
\else
\language=\csname l@#1\endcsname
\fi
#2}}
\providecommand{\BIBdecl}{\relax}
\BIBdecl

\bibitem{gong:pami13}
Y.~Gong, S.~Lazebnik, A.~Gordo, and F.~Perronnin, ``Iterative quantization: {A}
  procrustean approach to learning binary codes for large-scale image
  retrieval,'' \emph{{IEEE} Trans. Pattern Anal. Mach. Intell.}, vol.~35,
  no.~12, pp. 2916--2929, 2013.

\bibitem{he:kdd10}
J.~He, W.~Liu, and S.~Chang, ``Scalable similarity search with optimized kernel
  hashing,'' in \emph{ACM SIGKDD}, 2010, pp. 1129--1138.

\bibitem{kulis:pami12}
B.~Kulis and K.~Grauman, ``Kernelized locality-sensitive hashing,''
  \emph{{IEEE} Trans. Pattern Anal. Mach. Intell.}, vol.~34, no.~6, pp.
  1092--1104, 2012.

\bibitem{andoni:focs06}
A.~Andoni and P.~Indyk, ``Near-optimal hashing algorithms for approximate
  nearest neighbor in high dimensions,'' in \emph{FOCS}, 2006, pp. 459--468.

\bibitem{krizhevsky:nips12}
A.~Krizhevsky, I.~Sutskever, and G.~E. Hinton, ``{ImageNet} classification with
  deep convolutional neural networks,'' in \emph{NIPS}, 2012, pp. 1106--1114.

\bibitem{simonyan:iclr15}
K.~Simonyan and A.~Zisserman, ``Very deep convolutional networks for
  large-scale image recognition,'' in \emph{ICLR}, 2015.

\bibitem{szegedy:cvpr15}
C.~Szegedy, W.~Liu, Y.~Jia, P.~Sermanet, S.~Reed, D.~Anguelov, D.~Erhan,
  V.~Vanhoucke, and A.~Rabinovich, ``Going deeper with convolutions,'' in
  \emph{CVPR}, 2015, pp. 1--9.

\bibitem{girshick:cvpr14}
R.~B. Girshick, J.~Donahue, T.~Darrell, and J.~Malik, ``Rich feature
  hierarchies for accurate object detection and semantic segmentation,'' in
  \emph{CVPR}, 2014, pp. 580--587.

\bibitem{sermanet:iclr14}
P.~Sermanet, D.~Eigen, X.~Zhang, M.~Mathieu, R.~Fergus, and Y.~LeCun,
  ``Overfeat: Integrated recognition, localization and detection using
  convolutional networks,'' in \emph{ICLR}, 2014.

\bibitem{long:cvpr15}
J.~Long, E.~Shelhamer, and T.~Darrell, ``Fully convolutional networks for
  semantic segmentation,'' in \emph{CVPR}, 2015, pp. 3431--3440.

\bibitem{russakovsky:ijcv15}
O.~Russakovsky, J.~Deng, H.~Su, J.~Krause, S.~Satheesh, S.~Ma, Z.~Huang,
  A.~Karpathy, A.~Khosla, M.~Bernstein, A.~C. Berg, and L.~Fei-Fei, ``{ImageNet
  Large Scale Visual Recognition Challenge},'' \emph{Int'l J. Computer
  Visionl}, pp. 211--252, 2015.

\bibitem{donahue:icml14}
J.~Donahue, Y.~Jia, O.~Vinyals, J.~Hoffman, N.~Zhang, E.~Tzeng, and T.~Darrell,
  ``{DeCAF}: {A} deep convolutional activation feature for generic visual
  recognition,'' in \emph{ICML}, 2014, pp. 647--655.

\bibitem{oquab:cvpr14}
M.~Oquab, L.~Bottou, I.~Laptev, and J.~Sivic, ``Learning and transferring
  mid-level image representations using convolutional neural networks,'' in
  \emph{CVPR}, 2014, pp. 1717--1724.

\bibitem{razavian:cvprws14}
A.~S. Razavian, H.~Azizpour, J.~Sullivan, and S.~Carlsson, ``{CNN} features
  off-the-shelf: An astounding baseline for recognition,'' in \emph{CVPR
  Workshops on DeepVision}, 2014.

\bibitem{babenko:eccv14}
A.~Babenko, A.~Slesarev, A.~Chigorin, and V.~S. Lempitsky, ``Neural codes for
  image retrieval,'' in \emph{ECCV}, 2014, pp. 584--599.

\bibitem{oliva:ijcv01}
A.~Oliva and A.~Torralba, ``Modeling the shape of the scene: {A} holistic
  representation of the spatial envelope,'' \emph{Int'l J. Computer Vision},
  vol.~42, no.~3, pp. 145--175, 2001.

\bibitem{dalal:cvpr05}
N.~Dalal and B.~Triggs, ``Histograms of oriented gradients for human
  detection,'' in \emph{CVPR}, 2005, pp. 886--893.

\bibitem{everingham:ijcv10}
M.~Everingham, L.~{Van~Gool}, C.~K.~I. Williams, J.~Winn, and A.~Zisserman,
  ``The {Pascal} {V}isual {O}bject {C}lasses ({VOC}) challenge,'' \emph{Int'l
  J. Computer Vision}, vol.~88, no.~2, pp. 303--338, 2010.

\bibitem{fei-fei:cvprw04}
L.~Fei-Fei, R.~Fergus, and P.~Perona, ``Learning generative visual models from
  few training examples: an incremental bayesian approach tested on 101 object
  categories,'' in \emph{CVPRW on Generative-Model Based Vision}, 2004.

\bibitem{philbin:cvpr07}
J.~Philbin, O.~Chum, M.~Isard, J.~Sivic, and A.~Zisserman, ``Object retrieval
  with large vocabularies and fast spatial matching,'' in \emph{CVPR}, 2007.

\bibitem{girshick:pami16}
R.~B. Girshick, J.~Donahue, T.~Darrell, and J.~Malik, ``Region-based
  convolutional networks for accurate object detection and segmentation,''
  \emph{{IEEE} Trans. Pattern Anal. Mach. Intell.}, vol.~38, no.~1, pp.
  142--158, 2016.

\bibitem{chatfield:bmvc14}
K.~Chatfield, K.~Simonyan, A.~Vedaldi, and A.~Zisserman, ``Return of the devil
  in the details: Delving deep into convolutional nets,'' in \emph{BMVC}, 2014.

\bibitem{lin:cvprw15}
K.~Lin, H.-F. Yang, J.-H. Hsiao, and C.-S. Chen, ``Deep learning of binary hash
  codes for fast image retrieval,'' in \emph{CVPRW on DeepVision: Deep Learning
  in Computer Vision}, 2015, pp. 27--35.

\bibitem{lin:icmr15}
K.~Lin, H.-F. Yang, K.-H. Liu, J.-H. Hsiao, and C.-S. Chen, ``Rapid clothing
  retrieval via deep learning of binary codes and hierarchical search,'' in
  \emph{ICMR}, 2015, pp. 499--502.

\bibitem{raginsky:nips09}
M.~Raginsky and S.~Lazebnik, ``Locality-sensitive binary codes from
  shift-invariant kernels,'' in \emph{NIPS}, 2009, pp. 1509--1517.

\bibitem{liu:icml11}
W.~Liu, J.~Wang, S.~Kumar, and S.~Chang, ``Hashing with graphs,'' in
  \emph{ICML}, 2011, pp. 1--8.

\bibitem{weiss:nips08}
Y.~Weiss, A.~Torralba, and R.~Fergus, ``Spectral hashing,'' in \emph{NIPS},
  2008, pp. 1753--1760.

\bibitem{mu:cvpr10}
Y.~Mu, J.~Shen, and S.~Yan, ``Weakly-supervised hashing in kernel space,'' in
  \emph{CVPR}, 2010, pp. 3344--3351.

\bibitem{wang:pami12}
J.~Wang, S.~Kumar, and S.~Chang, ``Semi-supervised hashing for large-scale
  search,'' \emph{{IEEE} Trans. Pattern Anal. Mach. Intell.}, vol.~34, no.~12,
  pp. 2393--2406, 2012.

\bibitem{wang:eccv14}
Q.~Wang, L.~Si, and D.~Zhang, ``Learning to hash with partial tags: Exploring
  correlation between tags and hashing bits for large scale image retrieval,''
  in \emph{ECCV}, 2014, pp. 378--392.

\bibitem{kulis:nips09}
B.~Kulis and T.~Darrell, ``Learning to hash with binary reconstructive
  embeddings,'' in \emph{NIPS}, 2009, pp. 1042--1050.

\bibitem{lin:iccv13}
G.~Lin, C.~Shen, D.~Suter, and A.~van~den Hengel, ``A general two-step approach
  to learning-based hashing,'' in \emph{ICCV}, 2013, pp. 2552--2559.

\bibitem{lin:cvpr14}
G.~Lin, C.~Shen, Q.~Shi, A.~van~den Hengel, and D.~Suter, ``Fast supervised
  hashing with decision trees for high-dimensional data,'' in \emph{CVPR},
  2014, pp. 1971--1978.

\bibitem{liu:cvpr12}
W.~Liu, J.~Wang, R.~Ji, Y.~Jiang, and S.~Chang, ``Supervised hashing with
  kernels,'' in \emph{CVPR}, 2012, pp. 2074--2081.

\bibitem{norouzi:icml11}
M.~Norouzi and D.~J. Fleet, ``Minimal loss hashing for compact binary codes,''
  in \emph{ICML}, 2011, pp. 353--360.

\bibitem{norouzi:nips12}
M.~Norouzi, D.~J. Fleet, and R.~Salakhutdinov, ``Hamming distance metric
  learning,'' in \emph{NIPS}, 2012, pp. 1070--1078.

\bibitem{shen:cvpr15}
F.~Shen, C.~Shen, W.~Liu, and H.~T. Shen, ``Supervised discrete hashing,'' in
  \emph{CVPR}, 2015, pp. 37--45.

\bibitem{wang:iccv13}
J.~Wang, W.~Liu, A.~X. Sun, and Y.~Jiang, ``Learning hash codes with listwise
  supervision,'' in \emph{ICCV}, 2013, pp. 3032--3039.

\bibitem{krizhevsky:esann11}
A.~Krizhevsky and G.~E. Hinton, ``Using very deep autoencoders for
  content-based image retrieval,'' in \emph{ESANN}, 2011.

\bibitem{lai:cvpr15}
H.~Lai, Y.~Pan, Y.~Liu, and S.~Yan, ``Simultaneous feature learning and hash
  coding with deep neural networks,'' in \emph{CVPR}, 2015, pp. 3270--3278.

\bibitem{liong:cvpr15}
V.~E. Liong, J.~Lu, G.~Wang, P.~Moulin, and J.~Zhou, ``Deep hashing for compact
  binary codes learning,'' in \emph{CVPR}, 2015, pp. 2475--2483.

\bibitem{salakhutdinov:ijar09}
R.~Salakhutdinov and G.~E. Hinton, ``Semantic hashing,'' \emph{Int. J. Approx.
  Reasoning}, vol.~50, no.~7, pp. 969--978, 2009.

\bibitem{xia:aaai14}
R.~Xia, Y.~Pan, H.~Lai, C.~Liu, and S.~Yan, ``Supervised hashing for image
  retreieval via image representation learning,'' in \emph{AAAI}, 2014, pp.
  2156--2162.

\bibitem{zhao:cvpr15}
F.~Zhao, Y.~Huang, L.~Wang, and T.~Tan, ``Deep semantic ranking based hashash
  for multi-label image retreieval,'' in \emph{CVPR}, 2015, pp. 1556--1564.

\bibitem{kang:icdm12}
Y.~Kang, S.~Kim, and S.~Choi, ``Deep learning to hash with multiple
  representations,'' in \emph{ICDM}, 2012, pp. 930--935.

\bibitem{hinto:neco06}
G.~E. Hinton, S.~Osindero, and Y.~W. Teh, ``A fast learning algorithm for deep
  belief nets,'' \emph{Neural Computation}, vol.~18, no.~7, pp. 1527--1554,
  2006.

\bibitem{pan:tkde10}
S.~J. Pan and Q.~Yang, ``A survey on transfer learning,'' \emph{{IEEE} Trans.
  Knowl. Data Eng.}, vol.~22, no.~10, pp. 1345--1359, 2010.

\bibitem{deng:cvpr09}
J.~Deng, W.~Dong, R.~Socher, L.-J. Li, K.~Li, and L.~Fei-Fei, ``Image{N}et: {A}
  large-scale hierarchical image database,'' in \emph{CVPR}, 2009, pp.
  248--255.

\bibitem{gong:eccv14}
Y.~Gong, L.~Wang, R.~Guo, and S.~Lazebnik, ``Multi-scale orderless pooling of
  deep convolutional activation features,'' in \emph{ECCV}, 2014, pp. 392--407.

\bibitem{babenko:iccv15}
A.~Babenko and V.~S. Lempitsky, ``Aggregating deep convolutional features for
  image retrieval,'' in \emph{ICCV}, 2015, pp. 1269--1277.

\bibitem{hsieh:icml08}
C.~Hsieh, K.~Chang, C.~Lin, S.~S. Keerthi, and S.~Sundararajan, ``A dual
  coordinate descent method for large-scale linear {SVM},'' in \emph{ICML},
  2008, pp. 408--415.

\bibitem{fan:jmlr08}
R.~Fan, K.~Chang, C.~Hsieh, X.~Wang, and C.~Lin, ``{LIBLINEAR:} {A} library for
  large linear classification,'' \emph{J. Machine Learning Research}, vol.~9,
  pp. 1871--1874, 2008.

\bibitem{jia:acmmm14}
Y.~Jia, E.~Shelhamer, J.~Donahue, S.~Karayev, J.~Long, R.~B. Girshick,
  S.~Guadarrama, and T.~Darrell, ``Caffe: Convolutional architecture for fast
  feature embedding,'' in \emph{ACM MM}, 2014, pp. 675--678.

\bibitem{krizhevsky:tr09}
A.~Krizhevsky, ``Learning multiple layers of features from tiny images,''
  \emph{Computer Science Department, University of Toronto, Tech. Report},
  2009.

\bibitem{chua:civr09}
T.-S. Chua, J.~Tang, R.~Hong, H.~Li, Z.~Luo, and Y.-T. Zheng, ``{NUS-WIDE}: {A}
  real-world web image database from {N}ational {U}niversity of {S}ingapore,''
  in \emph{ACM CIVR}, 2009.

\bibitem{xiao:cvpr10}
J.~Xiao, J.~Hays, K.~A. Ehinger, A.~Oliva, and A.~Torralba, ``{SUN} database:
  Large-scale scene recognition from abbey to zoo,'' in \emph{CVPR}, 2010, pp.
  3485--3492.

\bibitem{yu:cvpr14}
A.~Yu and K.~Grauman, ``Fine-grained visual comparisons with local learning,''
  in \emph{CVPR}, 2014, pp. 192--199.

\bibitem{philbin:cvpr08}
J.~Philbin, O.~Chum, M.~Isard, J.~Sivic, and A.~Zisserman, ``Lost in
  quantization: Improving particular object retrieval in large scale image
  databases,'' in \emph{CVPR}, 2008.

\bibitem{zhou:nips14}
B.~Zhou, {\`{A}}.~Lapedriza, J.~Xiao, A.~Torralba, and A.~Oliva, ``Learning
  deep features for scene recognition using places database,'' in \emph{NIPS},
  2014, pp. 487--495.

\bibitem{lin:iclr14}
M.~Lin, Q.~Chen, and S.~Yan, ``Network in network,'' in \emph{ICLR}, 2014.

\bibitem{he:cvpr16}
K.~He, X.~Zhang, S.~Ren, and J.~Sun, ``Deep residual learning for image
  recognition,'' in \emph{CVPR}, 2016.

\bibitem{morere:cbmm06}
O.~Mor{\`e}re, A.~Veillard, J.~Lin, J.~Petta, V.~Chandrasekhar, and T.~Poggio,
  ``Group invariant deep representations for image instance retrieval,''
  \emph{Center for Brains, Minds and Machines, MIT Report}, 2016.

\bibitem{ng:cvprws15}
J.~Y. Ng, F.~Yang, and L.~S. Davis, ``Exploiting local features from deep
  networks for image retrieval,'' in \emph{CVPR Workshops}, 2015, pp. 53--61.

\bibitem{zeiler:iclr13}
M.~D. Zeiler and R.~Fergus, ``Stochastic pooling for regularization of deep
  convolutional neural networks,'' in \emph{ICLR}, 2013.

\bibitem{snoek:nips12}
J.~Snoek, H.~Larochelle, and R.~P. Adams, ``Practical bayesian optimization of
  machine learning algorithms,'' in \emph{NIPS}, 2012, pp. 2960--2968.

\bibitem{jie:accv14}
Z.~Jie and S.~Yan, ``Robust scene classification with cross-level {LLC} coding
  on {CNN} features,'' in \emph{ACCV}, 2014, pp. 376--390.

\end{thebibliography}


%
%
%

%
%
%

%
\vspace*{-3.5em}
\begin{IEEEbiography}[{\includegraphics[width=1in,height=1.25in,clip,keepaspectratio]{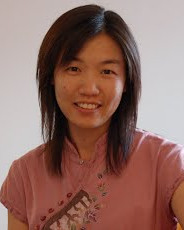}}]{Huei-Fang Yang} received the Ph.D. degree in computer science from Texas A\&M University, College Station, TX, USA, in 2011.
She is currently a post-doc researcher with the Research Center for Information Technology Innovation, Academia Sinica, Taiwan.
Her research interests include computer vision, machine learning, and biomedical image analysis.
\end{IEEEbiography}
\vspace*{-3.5em}
\begin{IEEEbiography}[{\includegraphics[width=1in,height=1.25in,clip,keepaspectratio]{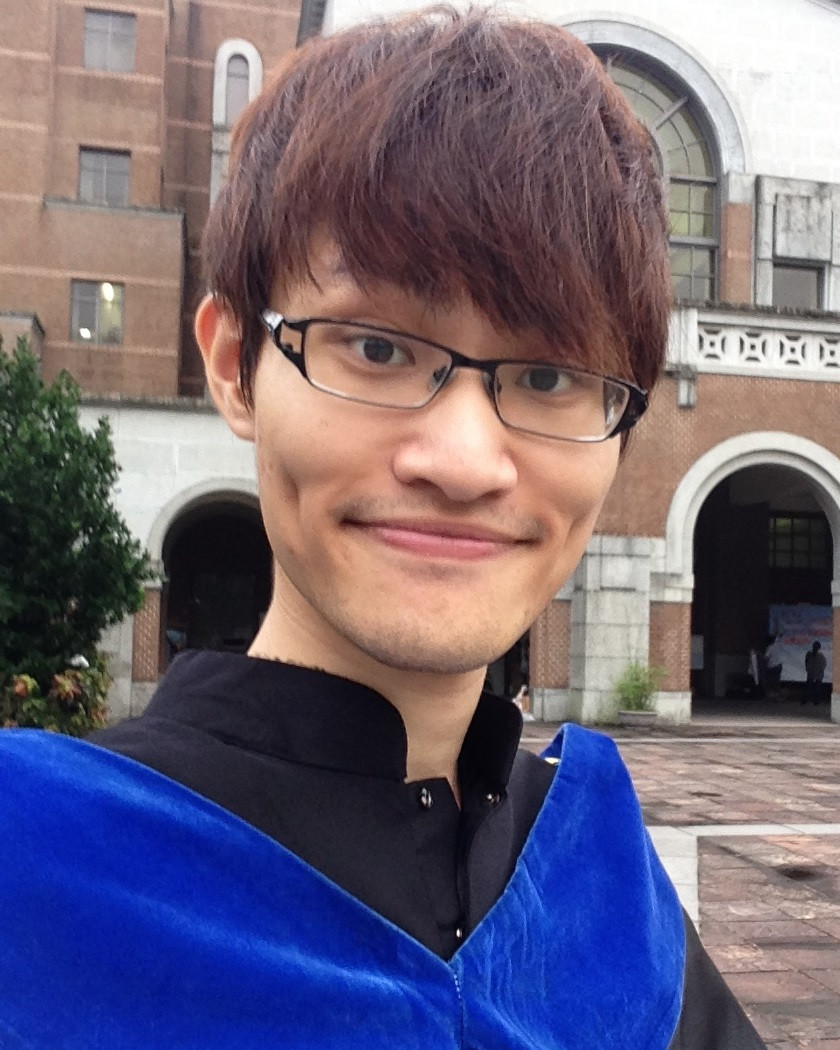}}]{Kevin Lin} received the M.S.~degree from the Graduate Institute of Networking and Multimedia, National Taiwan University, Taiwan, in 2014.
He is currently a Ph.D. student in the Dept.~Electrical Engineering, University of Washington, Seattle, WA.
Prior to his study, he was a Research Assistant with the Institute of Information Science, Academia Sinica, Taiwan.
His research interests include computer vision, machine learning, and multimedia.
\end{IEEEbiography}
\vspace*{-4em}
\begin{IEEEbiography}[{\includegraphics[width=1in,height=1.25in,clip,keepaspectratio]{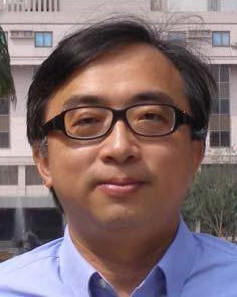}}]{Chu-Song Chen} is a Research Fellow with the Institute of Information Science, 
Academia Sinica, Taiwan.
His research interests include computer vision, image processing, pattern recognition, and multimedia.
He served as an Area Chair of ACCV '09 and ACCV'10, the Program Chair of IMV'12 and IMV'13, the Tutorial Chair of ACCV'14, the General Chair of IMEV'14, and the Workshop Chair of ACCV'16.
He is on the Editorial Board of the 
\textit{Machine Vision and Applications} journal. 
\end{IEEEbiography}
%

%
%
%

\newpage
\appendices

\section{Remark of classification results on ILSVRC}

In our experiments, the classification accuracies of SSDH and fine-tuned models are computed using only the center crop of a test image.
To have a fair comparison, we report the results of AlexNet and VGG on ILSVRC2012 based on a single crop.
Hence, there are discrepancies between our reported results and the ones in~\cite{krizhevsky:nips12,simonyan:iclr15} that employ multiple crops at test time.

In addition, because the top-5 accuracy is used to evaluate the algorithms in the ILSVRC competition, we report this accuracy for ILSVRC in Table~\ref{tbl:all-datasets-classification} as well.

It is worth noting that adding the latent layers does not necessarily reduce the classification accuracies.
We owe this to the following reason.
The added latent layer can also be interpreted as a dimension-reduction layer from the 4096-dimensional feature layer in AlexNet \textcolor{black}{or VGG}. 
Adding such a dimension-reduction layer is likely to remove the redundancy and achieve further performance gains for classification 
even when the latent layer outputs are restricted to be binary.
\end{document}